\documentclass[journal]{IEEEtran}
\ifCLASSOPTIONcompsoc
  \usepackage[nocompress]{cite}
\else
  \usepackage{cite}
\fi
\ifCLASSINFOpdf
\else
\fi
\usepackage{booktabs}
\usepackage{graphicx}
\usepackage{enumerate}
\usepackage{subfigure}
\usepackage{cite}

\usepackage{algorithm}
\usepackage{algorithmic}
\usepackage{graphics}

\usepackage{indentfirst}
\usepackage{verbatim}

\usepackage{amssymb}
\usepackage{longtable}
\usepackage{epsfig} 
\usepackage{amsthm}
\usepackage{mathrsfs}
\usepackage{amsmath}
\usepackage{stfloats}
\usepackage{multirow}
\usepackage{caption}
\usepackage{makecell}
\usepackage[hang,flushmargin]{footmisc}

\newtheorem{myDef}{Definition}

\begin{document}
%
\title{Multi-Party Dual Learning}
%
%
%
%

\author{Maoguo~Gong,~\IEEEmembership{Senior Member,~IEEE,}
  Yuan~Gao,
  Yu~Xie,
  A.~K.~Qin,~\IEEEmembership{Senior Member,~IEEE,}\\
  Ke~Pan,
  and~Yew-Soon Ong,~\IEEEmembership{Fellow,~IEEE,}
\thanks{Maoguo Gong, Yuan Gao, and Ke Pan are with the School of Electronic Engineering, Key Laboratory of Intelligent Perception and Image Understanding of Ministry of Education, Xidian University, Xi'an, Shaanxi Province 710071, China.
(e-mail: gong@ieee.org; cn\_gaoyuan@foxmail.com; kpansxxa@gmail.com)}
\thanks{Yu Xie is with the Key Laboratory of Computational Intelligence and Chinese Information Processing of Ministry of Education, Shanxi University, Taiyuan 030006, China. (e-mail: sxlljcxy@gmail.com)}
\thanks{A. K. Qin is with the Department of Computer Science and Software Engineering, Swinburne University of Technology, Melbourne, Australia. (e-mail: kqin@swin.edu.au)}
\thanks{Yew-Soon Ong is with the School of Computer Science and Engineering, Nanyang Technological University, Singapore 639798 (e-mail: asysong@ntu.edu.sg)}
}

%
%

\markboth{}%
{Shell \MakeLowercase{\textit{et al.}}: Bare Demo of IEEEtran.cls for IEEE Journals}
%



%
\maketitle
\begin{abstract}
The performance of machine learning algorithms heavily relies on the availability of a large amount of training data. However, in reality, data usually reside in distributed parties such as different institutions and may not be directly gathered and integrated due to various data policy constraints. As a result, some parties may suffer from insufficient data available for training machine learning models. In this paper, we propose a multi-party dual learning (MPDL) framework to alleviate the problem of limited data with poor quality in an isolated party. Since the knowledge sharing processes for multiple parties always emerge in dual forms, we show that dual learning is naturally suitable to handle the challenge of missing data, and explicitly exploits the probabilistic correlation and structural relationship between dual tasks to regularize the training process. We introduce a feature-oriented differential privacy with mathematical proof, in order to avoid possible privacy leakage of raw features in the dual inference process. The approach requires minimal modifications to the existing multi-party learning structure, and each party can build flexible and powerful models separately, whose accuracy is no less than non-distributed self-learning approaches. The MPDL framework achieves significant improvement compared with state-of-the-art multi-party learning methods, as we demonstrated through simulations on real-world datasets.

\end{abstract}

\begin{IEEEkeywords}
Multi-party learning, dual learning, privacy preservation.
\end{IEEEkeywords}



%
\IEEEpeerreviewmaketitle

\section{Introduction}\label{para:1}

%
%
%
%

\IEEEPARstart{D}{ata} is the oil for the operation of artificial intelligence. Nevertheless, with the exception of a few industries, the data available in most fields are of limited quantity or poor quality, making it hard to realize effective artificial intelligence applications. Specifically, many institutions may only have unlabeled data, while some others hold a limited amount of labeled data, and data exist in the form of isolated islands. Due to various data policy restrictions, it is almost impossible to integrate the data scattered around different institutions. For example, General Data Protection Regulation (GDPR) enforced by the European Union attaches importance to data privacy and user security; China's Cybersecurity Law enacted recently requires that Internet businesses must ensure the protection of personal information; California Consumer Privacy Act (CCPA), which became effective in 2020, also created new consumer rights relating to the access to personal information collected by businesses. Due to the above data regulations and laws, it is very difficult in many situations to break the barriers between data sources, since we are forbidden to collect or fuse data in different parties for machine learning tasks. Thus, it has been increasingly challenging to build efficient joint models while meeting privacy, security and regulatory requirements, especially with scattered data or limited labels.

In the past few years, there has been a growing interest in the privacy-preserving multi-party learning framework for addressing the challenge. A preliminary investigation is given in \cite{bonawitz2017practical}, where a global model is updated by distributed parties while keeping their data locally. It focuses on on-device multi-party learning that involves distributed mobile user interactions, and users' local parameters are uploaded with secure aggregation to update the global model. To avoid potential risks associated with gradient leakage, a differentially private regression analysis model based on relevance was presented in \cite{gong2019differential}. Yang \emph{et al.} \cite{yang2019federated} introduced various multi-party learning frameworks, which are classified based on how data are distributed among various participants in the sample ID and feature space, extending the concept of multi-party learning to cover collaborative learning scenarios among institutions. Gong~\emph{et~al.} \cite{gong2020privacy} integrated differential privacy and homomorphic encryption into the multi-party deep learning framework to prevent potential privacy leakage to other participants and the central server, and the framework works without requiring a manager that all participants trust. Moreover, there are many learning paradigms extending their models to consider the privacy requirement, such as reinforcement learning \cite{zhuo2019federated} and multi-task learning \cite{smith2017federated}. In general, these distributed learning approaches for a two-party scenario consist of two parts, which are encrypted entity alignment and secure model training, and the machine learning model will be trained on the common entities.

\begin{figure*}[htb!]
  \centering
  \subfigure[Federated Transfer Learning]{
    \includegraphics[width=0.4\linewidth]{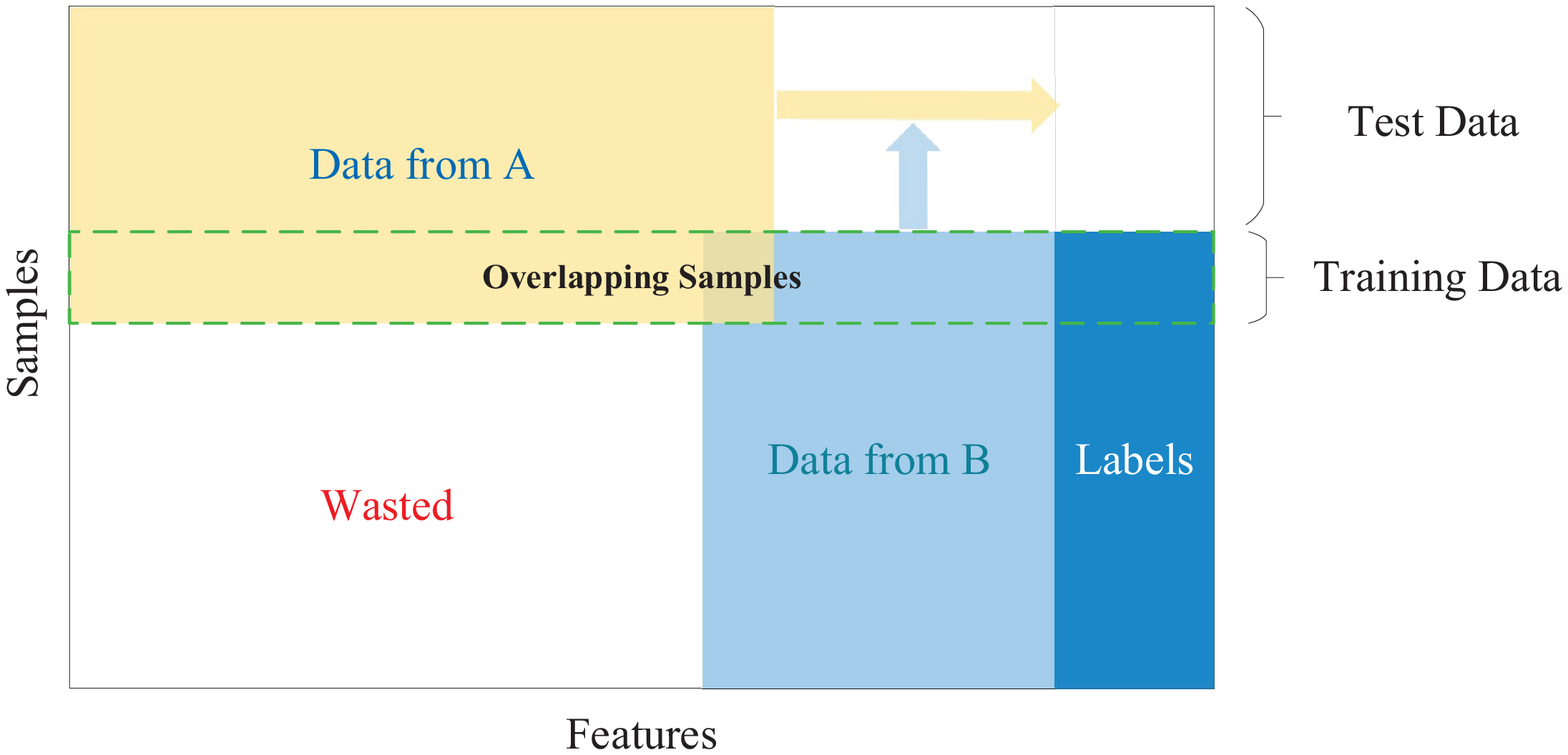}
    \label{fig:tab_w}}
  \subfigure[Multi-Party Dual Learning]{
    \includegraphics[width=0.4\linewidth]{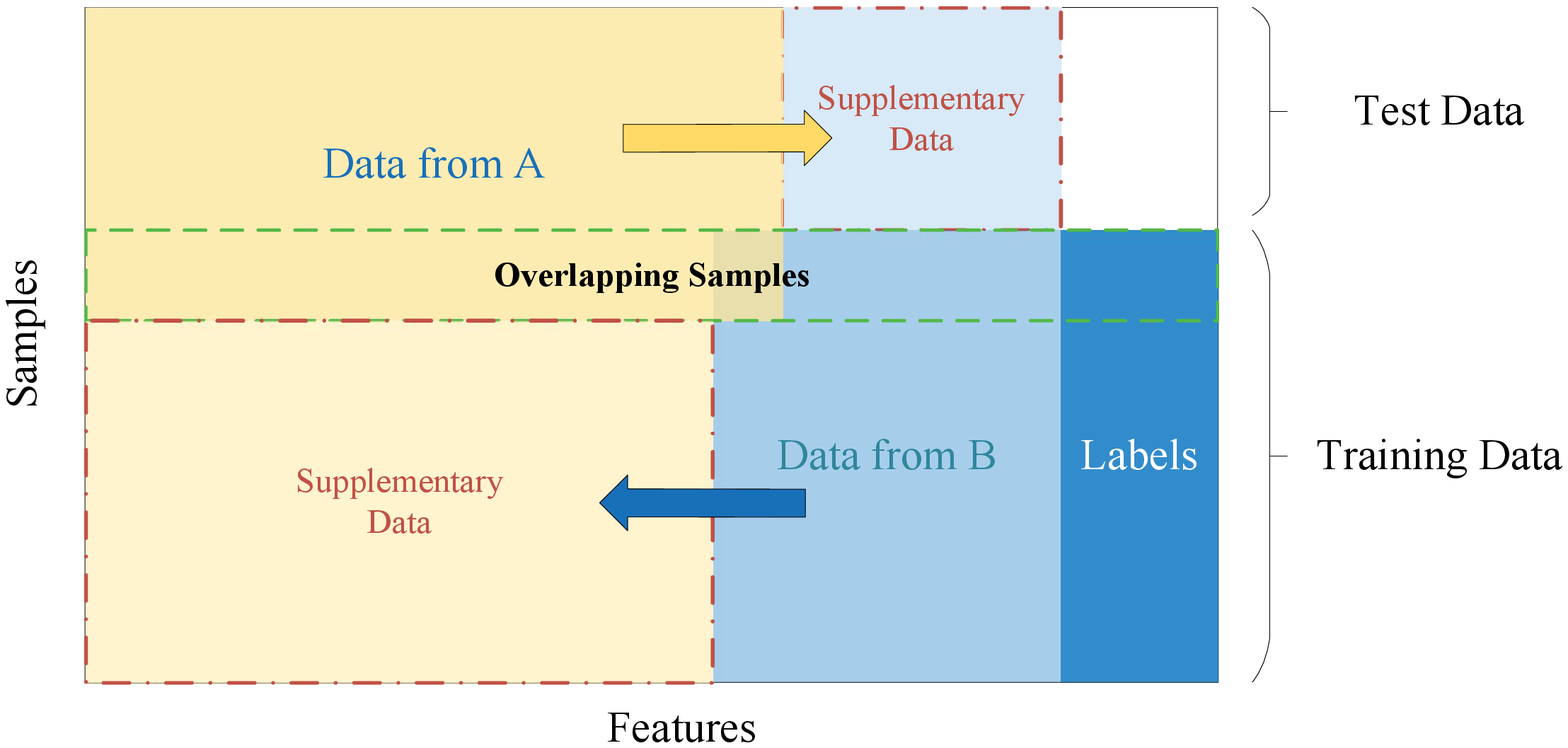}
    \label{fig:tab}}
  \caption{Comparision of models in the sample and feature space for a two-party learning problem.}
  \label{fig:1}
\end{figure*}

Most existing approaches are merely applicable to common features or samples under a secure framework \cite{hua2018collaborative, jayaraman2018distributed, huang2020dp, pillonetto2019distributed}. Nevertheless, for most real-world enterprises, as their business scope could be entirely different, the set of common entities and features could be small, leaving the majority of non-overlapping data wasted. To instantiate the situation in Fig.~\ref{fig:1}, consider two institutions, a bank and an e-commerce company located in different cities. Suppose both parties want to collaborate to build a joint prediction model for product purchase based on product and user information. However, only a small portion of the feature space in both parties overlaps because of different businesses. Besides, the user groups of them may have a small intersection due to the geographical restriction, which makes it hard to jointly build an effective model with data from both parties collaboratively. 

Federated transfer learning (FTL) \cite{liu2018secure} leverages transfer learning techniques \cite{pan2009survey} to capture the common knowledge of data from different sources. Overlapping data of the two parties are mapped to the same domain, and the distance between them is minimized so that the shared knowledge is able to be applied for subsequent machine learning tasks. With complementary knowledge transferred in the network, each party is able to run the entire model independently without features from the other party. Federated transfer learning extracts common features of data from both participants for downstream applications; however, their unique characteristics are undermined, especially when the correlation between two domains is relatively weak and only a few common features exist. Besides, since federated transfer learning only utilizes a limited set of co-occurrence samples, the value of the remaining precious labeled data is not exploited, as Fig.~\ref{fig:tab_w} shows.

To address the aforementioned challenges, we present a novel multi-party dual learning (MPDL) framework for missing data completion and label prediction. Since knowledge sharing for multiple parties is a mutual process, dual learning is introduced to strengthen the intrinsic probabilistic connection between the two parts of data. The dual generative models are trained on overlapping samples, and they are able to infer perturbed features from the other party. We introduce a feature-oriented differential privacy and deploy an affine transformation layer for it as a preprocessing step, thus dual training and testing are conducted without privacy disclosure. The dual inference is a general approach for data supplement, where the privacy-preserving setting relaxes the restrictions on downstream vertical learning algorithms, which could fully utilize all labeled data, as shown in Fig.~\ref{fig:tab}. Specifically, the main contributions of this paper are as follows: 
\begin{itemize}
\item{We introduce multi-party dual learning to provide solutions for multi-party learning problems, and we show that it is a natural choice for handling the challenge of insufficient overlapping pairs in the multi-party setting.}
\item{We provide a novel approach for leveraging labeled data in participants. The dual learning process is capable of expanding non-overlapping samples to overlapping pairs without privacy disclosure of the raw data, as a specially designed feature-oriented differential privacy is introduced to preserve private features.}
\item{The MPDL framework is extensible to various effective models, since the data structure is almost lossless. Experimental results demonstrate that the MPDL framework has superior performance over the non-distributed learning with gathered data.}
\end{itemize}

The remainder of this paper is organized as follows. Section~\ref{para:2} briefly presents the related backgrounds about multi-party learning, dual learning and privacy-preserving techniques. In Section~\ref{para:3}, we give a formal definition of the multi-party learning problem, then the details of our framework are described. Section~\ref{para:4} shows extensive experiments to validate the effectiveness. Finally, we conclude with a discussion of our framework and summarize the future work in Section~\ref{para:5}.

\section{Background and Related Works}
\label{para:2}

\subsection{Multi-Party Learning}
Multi-party learning is a general concept for all distributed collaborative machine learning techniques. For a multi-party learning problem, the data held by each party $i$ is denoted as $\mathcal{D}_i$. Specifically, we denote features space as $\mathcal{X}$, label space as $\mathcal{Y}$ and sample ID space as $\mathcal{I}$, and they constitute the complete training set $(\mathcal{I}_i,\mathcal{X}_i,\mathcal{Y}_i)$. The feature and sample space of the data owners could not be identical, and multi-party learning is classified based on how data is distributed among various parties in two spaces \cite{yang2019federated}.
\subsubsection{Sample-Based Learning}
It is applicable to the case that $\mathcal{D}_i$ shares the same feature space but different sample space for each party, and the symbolic representation can be summarized as:
\begin{equation}
\mathcal{X}_i=\mathcal{X}_j,\mathcal{Y}_i=\mathcal{Y}_j,\mathcal{I}_i\ne\mathcal{I}_j,\forall\mathcal{D}_i,\mathcal{D}_j, i\ne j.
\end{equation}

Hao \emph{et al.} \cite{hao2019efficient} proposed a non-interactive approach that can prevent private data from being leaked even though multiple entities collude with each other. Shokri \emph{et al.} \cite{shokri2015privacy} presented a collaboratively deep learning solution where each participant is trained independently and uploads only subsets of parameters to the central server. Likewise, Bonawitz \emph{et al.} \cite{bonawitz2017practical} proposed a secure aggregation scheme to protect the privacy of aggregated user updates under a horizontal multi-party learning framework. Recently, Phong \emph{et al.} \cite{phong2018privacy} introduced additively homomorphic encryption for parameter aggregation to provide privacy protections against the server.
\subsubsection{Feature-Based Learning}
It is introduced in the scenarios that data from the two parties share the same sample space but differ in feature space, and we have:
\begin{equation}
\mathcal{X}_i\ne\mathcal{X}_j,\mathcal{Y}_i\ne\mathcal{Y}_j,\mathcal{I}_i=\mathcal{I}_j,\forall\mathcal{D}_i,\mathcal{D}_j, i\ne j,
\end{equation}
in which the label $\mathcal{Y}$ may be held by only one party. In the learning process, different features are aggregated in a privacy-preserving manner. With the external supplementary features from the other party, a more effective model is built cooperatively. To facilitate the secure computations of the central model, a third-party collaborator is introduced, and the loss and gradients can be transferred losslessly during the training phase \cite{hardy2017private}. Therefore the model could reach the same level of accuracy as the non-distributed model using co-occurrence samples.
\subsubsection{Mapping-Based Learning}
It is appropriate for the case that two datasets differ in both sample ID space and feature space:
\begin{equation}
\mathcal{X}_i\ne\mathcal{X}_j,\mathcal{Y}_i\ne\mathcal{Y}_j,\mathcal{I}_i\ne\mathcal{I}_j,\forall\mathcal{D}_i,\mathcal{D}_j, i\ne j.
\end{equation}

In the scenario, only a small portion of the user group from the two parties overlaps, and their feature space $\mathcal{X}$ has a small or even no intersection. To handle this situation, transfer learning techniques are introduced and features of the two parties are mapped into the same low-dimensional space using the limited co-occurrence samples \cite{liu2018secure}. The common representation is applied to the subsequent machine learning tasks with only one-side features, thus the model is able to work for non-overlapping samples.

However, the mapped representations lead to the problem of structure and information loss compared with raw data. Moreover, data in the source-domain party (B in the case) that can be leveraged for the training process are wasted, as shown in Fig.~\ref{fig:tab_w}, making the central model overfitting on co-occurrence samples. To overcome the challenges, we introduce a dual learning scheme for the two-party learning problem, where both datasets differ in sample and feature space.

\subsection{Dual Learning}
Many supervised learning tasks emerge in dual forms. An example is machine translation, where millions of bilingual sentence pairs are needed for training. It is incredibly time-consuming and impractical to label them manually. However, the primal and dual tasks form a closed loop and provide informative feedback signals to train the models, even if without the involvement of a human labeler \cite{xia2017dual}. Based on the generation likelihood of the output of a model, and the reconstruction error of the raw data after the primal and dual processing, dual models could achieve a comparable accuracy with a small part of data to models trained from the full data. Dual learning has demonstrated its effectiveness in neural machine translation \cite{he2016dual}, image processing \cite{yi2017dualgan}, semantic segmentation \cite{luo2017deep} and sentiment analysis \cite{xia2018model}.

A dual learning scheme involves a primal task and its dual task. The primal task maps a sample from space $\mathcal{X}_A$ to space $\mathcal{X}_B$, and the dual task performs the reverse process. Specifically, dual models learn the conditional distribution $P(x^B|x^A;\theta_{AB})$ and $P(x^A|x^B;\theta_{BA})$ separately, where $x^A \in \mathcal{X}_A$ and $x^B \in \mathcal{X}_B$. The dual tasks are jointly learned, and their probabilistic correlation is exploited to improve the learning effectiveness. Ideally, we should have the probabilistic duality
\begin{equation}
\begin{split}
P(x^A)P(x^B|x^A;& \theta_{AB})=P(x^B)P(x^A|x^B;\theta_{BA})\\
& =P(x^A,x^B),
\end{split}
\label{equ:prob}
\end{equation}
where $P(x^A)$ and $P(x^B)$ are the marginal distributions, and they serve as the guarantee for the optimality of the dual models. 

In a two-party learning process, how to fully utilize the co-occurrence pairs and labeled samples so as to improve the central model remains a challenge. The dual learning mechanism greatly reduces the reliance on labeled data and provides a new perspective on leveraging misaligned data to train models. In order to make prediction while facing non-overlapping samples, the missing data are supposed to be inferred from the corresponding samples of the other party, thus the original information and structure are still lossless. Observing the existence of structure duality among the two models, we propose an secure multi-party dual learning framework.

\subsection{Privacy Preservation}
Privacy is one of the essential properties in a multi-party setting, which requires models to provide meaningful privacy guarantees. There have been various privacy-preserving techniques proposed in there years, e.g. homomorphic encryption \cite{phong2018privacy} differential privacy \cite{wang2019collecting}, and functional encryption \cite{shanmugam2018adaptive}. Besides, secure computing protocols and environments have been studied recently as well, such as secure multi-party computation \cite{zhao2019secure} and execution in trusted hardware \cite{8894393}. 

Among them, the methods of differential privacy are mature and well-accepted, which take the form of injecting noise to the data or parameters until the adversary cannot distinguish the sensitive individuals, thereby the user privacy can be protected. In order to prevent information leakage in the uploaded shared models in sample-based learning, Wei~\emph{et al.} \cite{wei2020federated} developed a novel privacy-preserving framework based on the concept of differential privacy. They also derived the theoretical convergence upper-bound of the algorithm and obtained the optimal number of communication rounds, which effectively improve the training efficiency for a given privacy level \cite{wei2020performance}. Since the injected noise in raw data could lead in decreased accuracy of the model, these differential privacy approaches involve a trade-off between privacy and accuracy.

\begin{figure*}[!ht]
  \centering
  \includegraphics[width=0.66\linewidth]{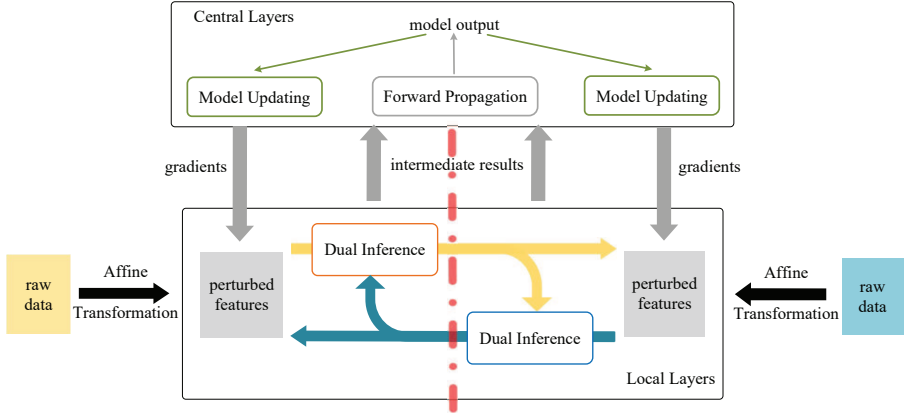}
\caption{Architecture of the Multi-Party Dual Learning scheme. Inferred data are produced via two local generative models (dual models) and aggregated with perturbed data from each party, which are transmitted to the collaborator for the training of the central model.}
\label{fig:frame}
\end{figure*}

\begin{myDef}
{\rm ($\epsilon$-Differential Privacy \cite{dwork2006calibrating}). Consider two neighboring datasets $\mathcal{D}_1$ and $\mathcal{D}_2$, $\mathcal{M}$ is a randomized algorithm and $\mathcal{O}$ means possible outputs, and we have}
\end{myDef}
\begin{equation}
Pr[\mathcal{M}(D_1)=\mathcal{O}]\leq e^{\epsilon}Pr[\mathcal{M}(D_2)=\mathcal{O}],
\end{equation}
where the privacy budget $\epsilon$ controls the probabilities that $\mathcal{M}$ obtains the same $\mathcal{O}$ on two neighboring datasets. It indicates the level of privacy preservation that $\mathcal{M}$ can provide, and a smaller $\epsilon$ enforces a stronger privacy guarantee of $\mathcal{M}$.
A general approach for meeting the requirements of differential privacy preservation of any function $\mathcal{F}$ on $\mathcal{D}$ is the Laplace mechanism \cite{dwork2006calibrating}, and it is achieved by injecting Laplace noise into the output of $\mathcal{F}$. The mechanism exploits the global sensitivity of $\mathcal{F}$ over any two neighboring datasets, which is a key parameter to determine the amount of noise injected in.

\begin{myDef}
{\rm (Global Sensitivity). Given the transformation function $\mathcal{F}$ on $\mathcal{D}$, the global sensitivity $GS_f(D)$ refers to the largest changes in the query results between the two neighboring datasets, which is defined as}
\end{myDef}
\begin{equation}
GS_f(D)={\max_{D_1,D_2}{||f(D_1)-f(D_2)||}}_1,
\end{equation}
where ${||f(D_1)-f(D_2)||}_1$ is the $L_1$ distance between $f(D_1)$ and $f(D_2)$.

\begin{myDef}
(Laplace Mechanism). Given the dataset $\mathcal{D}$, it is assumed that the global sensitivity of $\mathcal{F}$ is $\Delta$. The algorithm $\mathcal{M}(D)=f(D)+\eta$ provides $\epsilon$-differential privacy preservation, where $\eta$ is drawn i.i.d. from Laplace distribution with zero mean and scale, i.e.  $\eta\sim Lap(\Delta/\epsilon)$.
\end{myDef}
\begin{equation}
pdf(\eta)=\frac{\epsilon}{2GS_f(D)}\exp{(-|\eta|\frac{\epsilon}{GS_f(D)})}
\end{equation}
We adopt the Laplace mechanism as the privacy-preserving method in the proposed framework to prevent other participants from deriving the local raw features in the distributed scheme. The details of the perturbation in the multi-party dual learning and the theoretical analysis of sensitivities and error bounds of differential privacy are given in Section~\ref{para:3}.

\section{Proposed Algorithm}
\label{para:3}

The multi-party dual learning framework consists of local dual models and the central model, as shown in Fig.~\ref{fig:frame}. Dual models learn the intrinsic probabilistic connection of overlapping samples and infer the missing data in each party, and the central model can conduct various tasks on a large amount of supplementary data. In this section, we first give a formal definition of the learning problem and introduce the dual learning scheme in a multi-party setting, and the privacy-preserving techniques within it and the encrypted parameter interactions are elaborated in detail. Then the we present the collaborative process among participants and the third party, which is followed by an introduction of dual cross evaluation. Finally, the privacy and security analysis is presented.

\subsection{Problem Statement}
\label{sec:31}
Consider there are two participants $A$ and $B$ with datasets $\mathcal{D}_A=\{(x_i^A)\}_{i\in N_A}$ and $\mathcal{D}_B=\{(x_i^B,y_i^B)\}_{i\in N_B}$, where $y^B$ are labels only held by $B$. $\mathcal{D}_A$ and $\mathcal{D}_B$ are raw data of the two parties and not allowed to be exposed to each other. We assume that there exists a limited set of co-occurrence samples $\mathcal{D}_C=\{(x_i^A,x_i^B,y_i^B)\}_{i\in N_C}$, and $N_C$ is the non-empty intersection between $N_A$ and $N_B$, which can be found by secure entity alignment techniques \cite{scannapieco2007privacy} \cite{wang2017privacy}. Without losing generality, suppose all participants are honest-but-curious and non-colluding, which means that each of them operates in accordance with the predetermined processes, but is curious about the data of other participants. It is realistic since all participants are willing to collaborate for obtaining a model with higher accuracy than that trained on their own data; accordingly, they could be curious about the private information of others but do not undermine the training processes deliberately. Therefore, the main target of multi-party learning is to jointly build a central model for machine learning tasks. Meanwhile, the scheme should avoid the leaking of privacy information (partially) and accessing the other party's raw data and model structure (fully) in the process, in order to achieve a balance between accuracy and privacy. Given the above settings, the scheme should satisfy the following properties:
\begin{itemize}
\item{\textbf{Accuracy.} The accuracy of the multi-party learning scheme for predicting labels $y$ should be no less than that of the model simply trained on the aggregated co-occurrence samples, otherwise there is no incentive for participants to collaborate.}
\item{\textbf{Data privacy.} In the dual learning process, each party is not capable to derive information about a single feature in the other party with high confidence, and it cannot infer the raw data distribution as well.}
\item{\textbf{Computational security.} In the operation of the whole training and testing stages, raw data $x$ of each party must be kept locally. Each participant learns no information about the model deployed in the other party and has no access to the other's raw data.}
\end{itemize}

\subsection{Multi-Party Dual Learning}
\subsubsection{Dual Learning in a Multi-Party Setting}

For a two-party learning problem, the dual learning scheme is capable of strengthening the intrinsic probabilistic connection between the two datasets, which exists implicitly due to the relevance between the two participants. Specifically, the primal task is defined to find a mapping function $f:x^A\mapsto x^B$ and the dual task to find a mapping function $g:x^B\mapsto x^A$ so that the predictions are similar to the real counterparts. A common practice to train $(f,g)$ is minimizing the empirical risk in space $\mathcal{X}_A$ and $\mathcal{X}_B$:
\begin{equation}
\begin{split}
{\rm min_{\theta_{AB}}}\ \frac{1}{N_C}\sum_{i\in N_C}\ell_{align}\,(f\,(x_i^A;\theta_{AB}),x_i^B),\\
{\rm min_{\theta_{BA}}}\ \frac{1}{N_C}\sum_{i\in N_C}\ell_{align}\,(g\,(x_i^B;\theta_{BA}),x_i^A),
\end{split}
\label{equ:loss}
\end{equation}
where $\ell_{align}$ denotes the alignment loss function. However, probabilistic duality is not considered if the two models are learned independently and separately, and there is no guarantee that Eq.~\ref{equ:prob} will hold.

To tackle this problem, Eq.~\ref{equ:prob} is introduced into the above multi-objective optimization as a constraint, and converted to a regularization term by the method of Lagrange multipliers:
\begin{equation}
\begin{split}
\ell_{dual}&=({\rm log}\,P(x^A)-{\rm log}\,P(\hat{x}^A|x^B;\theta_{BA})\\
&+{\rm log}\,P(\hat{x}^B|x^A;\theta_{AB})-{\rm log}\,P(x^B))^2,
\end{split}
\end{equation}
in which $\hat{x}$ indicates the generated data, and we have $\hat{x}^A=g(x^B_i;\theta_{BA})$, $\hat{x}^B=f(x^A_i;\theta_{AB})$. Without prior knowledge of data distribution and structure, the marginal distributions can be calculated by the Kernel Density Estimation \cite{botev2010kernel}:
\begin{equation}
P(x)=\frac{1}{Nh^d}\sum_{i=1}^{N}K(\frac{x-x_i}{h}).
\end{equation}
It is a non-parametric method to estimate the probability density function based on a finite data sample, which takes the density value of the neighborhood as the density function at $x$. In the estimation, $d$ is the dimension of data, and $K(\cdot)$ is a multi-dimensional kernel, generally the product of $d$ one-dimensional Gaussian kernels, which are non-negative and conform to the probability density property. Bandwidth $h$ is a smoothing parameter and indicates the sample radius. It essentially requires $h$ trend to 0 for small bias, while too few sampling points can lead to large variance. As there is no reliable bandwidth selection method for multi-dimensional estimation, its value is set to $1.05\cdot N^{-1/5}$ empirically for the bias-variance tradeoff. Since there is merely a small set of co-occurrence samples $\mathcal{D}_C$, $N$ is set to $N_A$ and $N_B$ for the two participants, and the sampling error could be reduced using more non-overlapping samples. The dual models are learned by minimizing the weighted combination between the alignment loss Eq.~\ref{equ:loss} and the above penalty term:
\begin{equation}
\mathcal{L}=\ell_{align}\,(\hat{x},x)+\lambda \ell_{dual}(\hat{x}^A,\hat{x}^B,x^A,x^B).
\label{equ:wl}
\end{equation}

Local models for party $A$ and $B$ are private and their structure could be different. Provided the shape of data from the other party, they can build targeted and efficient models without consulting with the partner, such as CNNs for images and RNNs for texts. Similarly, it is applicable for the generation between different data types as well in cases where the central model is well designed. Each party calculates gradients for the output layer based on the weighted loss Eq.~\ref{equ:wl} and sends them to the partner, who then computes gradients for the other layers and leverages different optimizers, such as SGD \cite{zinkevich2010parallelized}, Adadelta \cite{zeiler2012adadelta} or RMSprop \cite{mukkamala2017variants}, to update parameters of the local model. Once dual models converge, each participant is able to provide predictions for a specific user that lacks corresponding data for the other party. Receiving the predictions, the other participant could extend the dataset and join the multi-party learning process.

\subsubsection{Multi-Party Learning with Collaborator}

\begin{figure*}[!tb]
  \centering
  \includegraphics[width=0.66\linewidth]{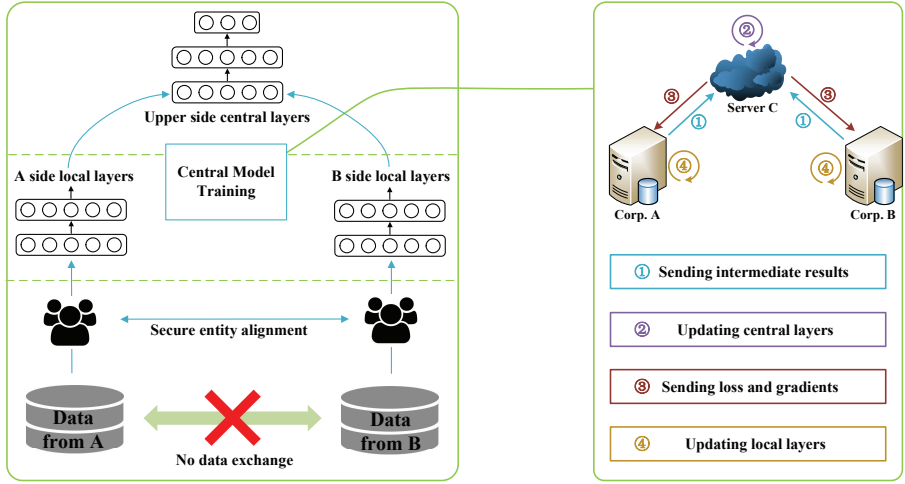}
\caption{Secure multi-party learning scheme with collaborator. Note that the supplementary data is assumed to have been inferred by dual models. The upper layers of the joint model is deployed in the central server, while the local layers are located in participants, and the model is updated by passing intermediate results and gradients. As each participant has no access to acquire information from the other party beyond what is revealed by the model output, the computational security is guaranteed in the process.}
\label{fig:thirdparty}
\end{figure*}
Different from the feature and sample setting in conventional distributed learning \cite{gong2020privacy} where data in all participants follow a similar distribution, the proposed framework focuses on solutions for unmatched features and fewer overlapping samples. Inspired by the feature-based learning schemes \cite{yang2019federated} \cite{hardy2017private} \cite{Chen2020VAFLAM}, which have been proved to be secure and lossless, a third-party collaborator $C$ is introduced for the training and testing of the central model that predicts labels for the joint learning task. Though it merely carries out parameter aggregation in sample-based algorithms \cite{gong2020privacy}, the collaborator shoulders more tasks such as model training and results generation in our feature-based approach. Note that it only assists the joint learning but do not participate in the dual learning process. 

Since the central model is deployed in $C$, we assume that the collaborator $C$ is honest and does not collude with party $A$ and $B$, otherwise the model parameters are directly exposed, making it possible to derive raw training data. It is a reasonable assumption as the collaborator can be played by authorities such as governments or the Trusted Execution Environment (TEE). Note that participants cannot upload their data to $C$ directly for data privacy and security reasons, even though $C$ is trusted or the data has been encrypted.

The training and inferring process of the central model is similar as that of a conventional vertical federated learning model, except that the input layers are located in different places (participants in this case), as shown in Fig.~\ref{fig:thirdparty}, and details of the process are given and analyzed in Section~\ref{sec:multi}. With the help of dual learning, the set of co-occurrence samples can be extended to $\mathcal{D}_C=\{(x_i^A,x_i^B,y_i^B)\}_{i\in N_B}$, which is sufficient enough for the training of the central model. However, the quality of the generated data is not yet guaranteed, thus the dual cross validation is introduced to improve the reliability of the data, see Algorithm~\ref{alg:mpdl}. 

\begin{algorithm}[!t]
  \caption{ Multi-Party Dual Learning}
  \begin{algorithmic}
  \label{alg:mpdl}
  \REQUIRE
  Data from both parties $\mathcal{D}_A$ and $\mathcal{D}_B$; threshold $T$; number of folds $K$; max iterations $m$
  \ENSURE
  Central model $\mathcal{M}_D$

  Split $\mathcal{D}_C$ into $K$ shares, $\mathcal{D}_C^k=\{(x_i^A,x_i^B)\}_{i\in I_k}$;\\
  Initialize dual models $\mathcal{M}_A$ and $\mathcal{M}_B$;
  \FOR{$j=1,2,...,m$}
      \STATE Pick a random $k\in [1,K]$;
      \STATE Reinitialize central models $\mathcal{M}_C$ and $\mathcal{M}_D$;
      \STATE Train dual models $\mathcal{M}_A$ and $\mathcal{M}_B$ on $\mathcal{D}_C$;
      \STATE Conduct prediction with $\mathcal{M}_B$ on $\{x_i^B\}_{i\in (N_B \setminus N_A)}$ to get $\mathcal{D}_P=\{(x_i^A,x_i^B)\}_{i\in (N_B \setminus N_A)}$;
    \STATE Train central model $\mathcal{M}_D$ and $\mathcal{M}_C$ on $\mathcal{D}_T^k \cup \mathcal{D}_P$ and $\mathcal{D}_T^k$, respectively;
    \STATE Evaluate $\mathcal{M}_C$ and $\mathcal{M}_D$ on $\mathcal{D}_V^k$ to get the performance of models $\mathcal{V}_C$ and $\mathcal{V}_D$;
    \IF{$\mathcal{V}_D-\mathcal{V}_C > T$}
       \STATE Break;
    \ENDIF
      
  \ENDFOR
  \RETURN $\mathcal{M}_D$
  \end{algorithmic}
\end{algorithm}
The co-occurrence samples $\mathcal{D}_C$ are split into K folds, which requires secure entity alignment techniques. Specifically, suppose party $B$ generates the public key and sends it to $A$. Party $A$ encrypts its sample IDs and masks them with a random number based on Hash algorithm, and sends the encrypted results to $B$. Then, party $B$ encrypts its sample IDs and further masks all encrypted sample IDs with a new random number introduced by Hash, and sends these encrypted sample IDs of both parties to $A$. Party $A$ could remove the mask of itself, find the encrypted intersection of the results and sends it to $B$ for decrypting. Afterwards, party $B$ obtains the co-occurrence sample IDs and specifies the division of K folds. In order to conduct dual validation, we reserve one of the K folds each time as the validation set $\mathcal{D}_V^k$, and the remaining data as training set $\mathcal{D}_T^k$.

\begin{figure*}[!ht]
  \centering
  \includegraphics[width=0.66\linewidth]{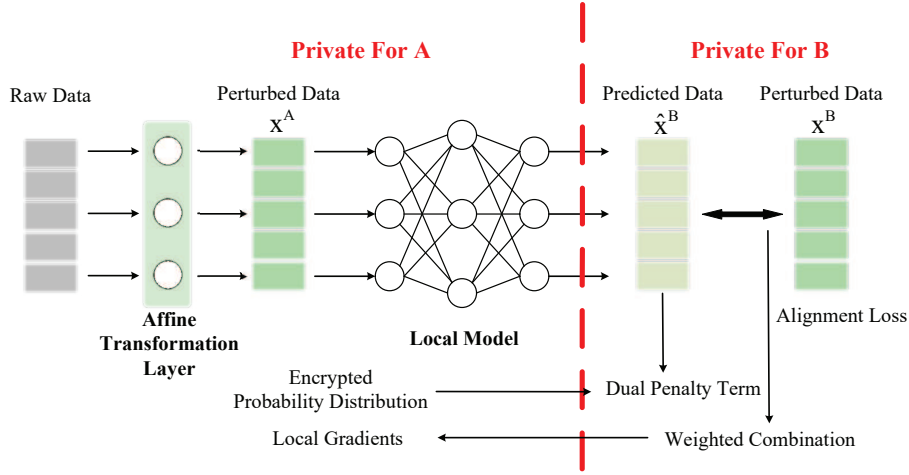}
\caption{A graphical illustration of the training process of the dual model in party $A$. After deriving the affine transformation layer assisted by the collaborator, raw data is perturbed with Laplace noise through the layer. Party $A$ sends encrypted probability distribution to party $B$ to calculate the regularization term. The weighted combination of the dual loss and the alignment loss are sent back to party $A$, who decrypts them and conducts the backpropagation to obtain local gradients.}
\label{fig:dual}
\end{figure*}
We utilize $\mathcal{D}_C$ to train dual models and conduct perturbed data prediction on $\{x_i^B\}_{i\in (N_B \setminus N_A)}$ to get $\mathcal{D}_P=\{(x_i^A,x_i^B)\}_{i\in (N_B \setminus N_A)}$. We denote $\mathcal{V}$ as the performance of the central model evaluated on the validation set $\mathcal{D}_V^k$, $\mathcal{V}_F$ as the model trained on $\mathcal{D}_T^k$, and $\mathcal{V}_D$ as the model trained on $\mathcal{D}_T^k \cup \mathcal{D}_P$. The dual models are considered to converge if $\mathcal{V}_D-\mathcal{V}_C > T$ is satisfied, where $T$ is a threshold, and we can get $\mathcal{D}_C=\{(x_i^A,x_i^B,y_i^B)\}_{i\in N_B}$ for the training of the central model. Once the above models are trained, we can provide predictions for unlabeled data in a participant. Specifically for each unlabeled data $x^A_u$, party $A$ sends the encrypted ID and the prediction $\hat{x}^B_u$ of model $\mathcal{M}_A$ to party $B$ and weighted input $\{z_j\}_A$ to collaborator $C$. If the overlapping pairs exist in party $B$, $\{z_j\}_B$ are calculated on $x^B_u$ and sent to $C$. Otherwise, party $B$ decrypts $\hat{x}^B_u$ and computes $\{z_j\}_B$ on it, then $C$ combines the result and sends it back to $A$.

\subsection{Secure and Privacy-Preserving MPDL}
\subsubsection{Feature-Oriented Differential Privacy}
Though the dual learning process is capable of expanding non-overlapping samples to overlapping pairs and improves the performance of the central model, it introduces a dilemma that the precise prediction of other party's raw data can lead to privacy disclosure. In ideal conditions, the dual models are able to accurately infer each party's data from the other, and there would be no privacy at all. However, data privacy needs to be preserved in this process, thereby raw data must be perturbed to avoid privacy leakage during the training. Therefore, there is a tradeoff between the accuracy of the central model and the privacy of the data.

A simple but effective method is to normalize the data and inject noise into it. The normalization process is conducive to the model convergence and makes it difficult for the other party to infer the raw data during the training without the prior knowledge of data distribution. Meanwhile, we preserve differential privacy of raw features in each party. Consider two neighboring datasets $\mathcal{D}_1$ and $\mathcal{D}_2$ differing at most one feature, $\mathcal{M}$ is the affine transformation and $\mathcal{O}$ means any possible output of $\mathcal{M}$, and it should satisfy

\begin{equation}
Pr[\mathcal{M}(D_1)=\mathcal{O}]\leq e^{\epsilon}Pr[\mathcal{M}(D_2)=\mathcal{O}],
\end{equation}

which is controlled by privacy budget $\epsilon$. Specifically, identical noise distribution $\frac{1}{|L|}Lap{\frac{\Delta_{\mathbf{h}_0}}{\epsilon}}$ to all input features to preserve differential privacy in the computation of $\textbf{h}_0$. Assume that two neighboring datasets with $L$ samples differ in the last feature, denoted as $\mathbf{x}_d$ ($\mathbf{x}_{d}'$) in the feature space $\mathcal{D}$ ($\mathcal{D}'$). The global sensitivity is
\begin{equation}
\begin{split}
&\Delta_{\mathbf{h}_0}=\sum_{h\in\mathbf{h}_0}\sum_{i=1}^L||\sum_{\mathbf{x}_j\in \mathcal{D}} x_{ij}-\sum_{\mathbf{x}_j'\in \mathcal{D}'} x_{ij}||_1 \\
&=\sum_{h\in\mathbf{h}_0}\sum_{i=1}^L||x_{id}-x_{id}'||_1 \le 2\max_{\mathbf{x}_j\in \mathcal{D}}\sum_{h\in\mathbf{h}_0}\sum_{i=1}^L||x_{ij}||_1.
\end{split}
\end{equation}
Since $\forall \mathbf{x}_{i,j}:x_{ij}\in[0,1]$, we have that $\Delta_{\mathbf{h}_0} \le2\sum\nolimits_{h\in\mathbf{h}_0} L$.

\emph{Lemma:} The $\epsilon$-differential privacy is preserved in the computation of $\textbf{h}_0$ for any two vertically split neighboring datasets differing at most one feature.

\emph{Proof:} Consider the static bias $b=1$ as the 0-th input feature for all samples and its associated parameter $W_b$, i.e., $x_{i0}=b=1$ and $W=W_b\cup W$, each neuron $h\in\textbf{h}_0$ can be re-written as:
\begin{equation}
\begin{split}
h_\mathcal{D}(W)&=\sum_{j=0}^d \left[\sum_{i=1}^L(x_{ij}+\frac{1}{|L|}Lap(\frac{\Delta_{\mathbf{h}_0}}{\epsilon}))W^T \right]\\
&=\sum_{j=0}^d \left[\sum_{i=1}^L x_{ij}+Lap(\frac{\Delta_{\mathbf{h}_0}}{\epsilon})W^T \right]=\sum_{j=0}^d\phi_j^h W^T,
\end{split}
\end{equation}
where $\phi_j^h=\left[\sum_{i=1}^L x_{ij}+Lap(\frac{\Delta_{\mathbf{h}_0}}{\epsilon}) \right]$.

We can see that $\phi_j^h$ is the perturbation of the input feature $x_{ij}$ associated with the j-th parameter $W_j\in W$ of the hidden neuron $h$ on a dataset. Since all hidden neurons $h\in\textbf{h}_0$ are perturbed, we have that:
\begin{equation}
Pr(\textbf{h}_{0\mathcal{D}}(W_0))=\prod_{h\in\textbf{h}_0}\prod_{j=0}^d \exp(\frac{\epsilon||\sum_{i=1}^L x_{ij}-\phi_j^h||}{\Delta_{\mathbf{h}_0}}),
\end{equation}
in which $\Delta_{\mathbf{h}_0}$ is set to $2\sum\nolimits_{h\in\mathbf{h}_0} L$ and $\textbf{h}_{0\mathcal{D}}(W_0)=\{h_\mathcal{D}(W)\}_{h\in\textbf{h}_0}$ is the output of the affine transformation layer. It is proved that:
\begin{equation}
\begin{split}
&\frac{Pr(\textbf{h}_{0\mathcal{D}}(W_0))}{Pr(\textbf{h}_{0\mathcal{D}'}(W_0))}=\frac{\prod_{h\in\textbf{h}_0}\prod_{j=0}^d \exp(\frac{\epsilon||\sum_{i=1}^L x_{ij}-\phi_j^h||}{\Delta_{\mathbf{h}_0}})}{\prod_{h\in\textbf{h}_0}\prod_{j=0}^{d'}\exp(\frac{\epsilon||\sum_{i=1}^L x_{ij}'-\phi_j^h||}{\Delta_{\mathbf{h}_0}})}\\
&= \frac{\prod_{h\in\textbf{h}_0}\prod_{j=0}^{d-1} \exp(\frac{\epsilon||\sum_{i=1}^L x_{ij}-\phi_j^h||}{\Delta_{\mathbf{h}_0}})\cdot\exp(\frac{\epsilon||\sum_{i=1}^L x_{id}-\phi_d^h||}{\Delta_{\mathbf{h}_0}})} {\prod_{h\in\textbf{h}_0}\prod_{j=0}^{d'-1}\exp(\frac{\epsilon||\sum_{i=1}^L x_{ij}'-\phi_j^h||}{\Delta_{\mathbf{h}_0}})\cdot\exp(\frac{\epsilon||\sum_{i=1}^L x_{id'}'-\phi_{d'}^h||}{\Delta_{\mathbf{h}_0}})} \\
&=\frac{\prod_{h\in\textbf{h}_0}\exp(\frac{\epsilon||\sum_{i=1}^L x_{id}-\phi_d^h||}{\Delta_{\mathbf{h}_0}})} {\prod_{h\in\textbf{h}_0}\exp(\frac{\epsilon||\sum_{i=1}^L x_{id'}'-\phi_{d'}^h||}{\Delta_{\mathbf{h}_0}})} \\
&\le \prod_{h\in\textbf{h}_0} \exp(\frac{\epsilon}{\Delta_{\mathbf{h}_0}}||\sum_{i=1}^L x_{id}-\sum_{i=1}^L x_{id'}'||_1)\\ 
&\le \prod_{h\in\textbf{h}_0}\prod_{i=1}^L \exp(\frac{2\epsilon}{\Delta_{\mathbf{h}_0}}\max_{\mathbf{x}_d\in \mathcal{D}}||x_{id}||_1)\le \prod_{h\in\textbf{h}_0}\prod_{i=1}^L \exp(\frac{2\epsilon}{\Delta_{\mathbf{h}_0}})\\
&\le\exp(\epsilon\cdot\frac{2\sum\nolimits_{h\in\mathbf{h}_0} L}{\Delta_{\mathbf{h}_0}})=\exp(\epsilon).
\end{split}
\end{equation}
Note that due to the identical noise distribution $\frac{1}{|L|}Lap(\frac{\Delta_{\mathbf{h}_0}}{\epsilon})$ injected into all input features, the same perturbation is held for the last feature $\mathbf{x}_d$ and $\mathbf{x}_{d}'$. Based on the above analysis, the $\epsilon$-differential privacy is preserved in the computation of $\textbf{h}_0$. After this preprocessing, the other hidden layers of the dual model are stacked on the top of the perturbed layer $\textbf{h}_0$; hence, all subsequent operations are based on the output of $\textbf{h}_0$ and have no direct access to the raw data. In the training of dual models, it becomes impossible for participants to distinguish the output of $\textbf{h}_0$ between $\mathcal{D}_1$ and $\mathcal{D}_2$ with different features. Each participant is not capable to infer private information about a single feature with high confidence in the other party, even if it knows all the remaining features in the dataset, thus data privacy can be preserved.

\subsubsection{Privacy-Preserving Dual Learning}

In the calculation of dual loss $\ell_{dual}$, which is a data-dependent regularization term and is the same for both models, information about data distribution $P(x)$ is private for each party and cannot be shared. Therefore, the encrypted probability distribution for a minibatch and gradients of the output layer needs to be computed and transmitted. The graphical illustration of the training process is shown in Fig.~\ref{fig:dual}. To be specific, due to the efficiency of the Paillier algorithm, it is employed as the additive homomorphic encryption algorithm \cite{acar2018survey} in the paper, and denoted as $[[\cdot]]_A$ and $[[\cdot]]_B$ for party $A$ and $B$, respectively. Note that the Paillier algorithm only supports the calculation of non-negative integers. Following the previous research in \cite{gong2020privacy}, the positive and negative floating-point numbers are encoded as non-negative numbers before encryption and decoded after calculation. 

On a minibatch of co-occurrence samples $\mathcal{D}_C$, the two parties produce inferred data $\hat{x}^B$ and $\hat{x}^A$ via their local neural networks independently. For the following steps, we take party $A$ as an example, and all the deduction can be adapted to party $B$. In addition, $P(\hat{x}^B|x^A;\theta_{AB})$ is expressed as $P(\hat{x}^B)$ for simplicity. The gradients of the dual loss $\ell_{dual}$ on the inferred data $\hat{x}^B$ is
\begin{equation}
\begin{split}
\nabla_{\hat{x}_B}\,\ell_{dual}=&\nabla_{\hat{x}_B} {\rm log}\,P(\hat{x}^B)({\rm log}\,P(\hat{x}^B)-{\rm log}\,P(x^B))\\
+&\nabla_{\hat{x}_B}{\rm log}\,P(\hat{x}^B)({\rm log}\,P(x^A)-{\rm log}\,P(\hat{x}^A)).
\label{equ:longeq}
\end{split}
\end{equation}

There have been a large number of works discussing the potential risks associated with gradient leakage \cite{shokri2015privacy} \cite{abadi2016deep}. To prevent party $B$ from cracking probability distribution that contains the knowledge of raw data distribution, $\hat{x}^B$ and encrypted components $[[{\rm log}\,P(x^A)-{\rm log}\,P(\hat{x}^A)]]_A$ are sent to party $B$ to assist calculations of $\ell_{align}\,(\hat{x}^B,x^B)$ and $\ell_{dual}(\hat{x}^A,\hat{x}^B,x^A,x^B)$. Party $B$ has the ground truth $x^B$ to compute the alignment loss, and it sends $\nabla_{\hat{x}_B}(\ell_{align}\,(\hat{x}^B,x^B))+\lambda\nabla_{\hat{x}_B} {\rm log}\,P(\hat{x}^B)({\rm log}\,P(\hat{x}^B)-{\rm log}\,P(x^B))$ and $\lambda\nabla_{\hat{x}_B}{\rm log}\,P(\hat{x}^B)[[{\rm log}\,P(x^A)-{\rm log}\,P(\hat{x}^A)]]_A$ back to party $A$. 
Note that the party $B$ is ignorant of the network structure of party $A$, so it can merely calculate gradients of the output layer $\nabla_{\hat{x}_B}$ rather than $\nabla_{\theta_{AB}}$. Then party $A$ decrypts the probability distribution and further computes gradients of each layer with the chain rule, and the back propagation could be carried out in the local model. The algorithm is shown in Algorithm~\ref{alg:dual}.

\begin{algorithm}[!t]
  \caption{ Privacy-Preserving Dual Learning}
  \begin{algorithmic}
  \label{alg:dual}
  \REQUIRE
  Lagrange parameters $\lambda_A$ and $\lambda_B$; co-occurrence samples $\mathcal{D}_C$; optimizers $Opt_A$ and $Opt_B$
  \ENSURE
  Dual model parameters $\theta_{AB}$ and $\theta_{BA}$

  $A$,$B$ initialize $\theta_{AB}$ and $\theta_{BA}$
  \REPEAT
      \STATE Get a minibatch of overlapping samples $\{(x^A_i,x^B_i)\}_{i=1}^m$

      \STATE \textbf{\emph{A}} do:\\
      Computes $\hat{x}_i^B\leftarrow f\,(x_i^A;\theta_{AB})$ for each $i$ and sends to \textbf{\emph{B}};\\

      \STATE \textbf{\emph{B}} do:\\
      Computes $\hat{x}_i^A\leftarrow f\,(x_i^B;\theta_{BA})$ for each $i$ and sends to \textbf{\emph{A}};\\
      Computes $\nabla_{\hat{x}_B}\sum\limits_{i=1}^{m}\ell(\hat{x}^B_i,x^B_i)+\lambda_B\sum\limits_{i=1}^{m}(\nabla_{\hat{x}_B}{\rm log}\,P(\hat{x}^B_i))\cdot$\\$({\rm log}\,P(\hat{x}^B_i)-{\rm log}\,P(x^B_i))$ and sends to \textbf{\emph{A}};\\
      Computes and encrypts $[[{\rm log}\,P(\hat{x}_i^B)-{\rm log}\,P(x_i^B)]]_B$ and sends to \textbf{\emph{A}};\\

      \STATE \textbf{\emph{A}} do:\\
      Computes $\nabla_{\hat{x}_A}\sum\limits_{i=1}^{m}\ell(\hat{x}^A_i,x^A_i)+\lambda_A\sum\limits_{i=1}^{m}(\nabla_{\hat{x}_A}{\rm log}\,P(\hat{x}^A_i))\cdot$\\$({\rm log}\,P(\hat{x}^A_i)-{\rm log}\,P(x^A_i))$ and sends to \textbf{\emph{B}};\\
      Computes and encrypts $[[{\rm log}\,P(\hat{x}_i^A)-{\rm log}\,P(x_i^A)]]_A$ and sends to \textbf{\emph{B}};\\
      Computes $\lambda_A\nabla_{\hat{x}_A}{\rm log}P(\hat{x}^A_i)\cdot[[{\rm log}P(x^B_i)-{\rm log}P(\hat{x}^B_i)]]_B$ and sends to \textbf{\emph{B}};\\

      \STATE \textbf{\emph{B}} do:\\
      Computes $\lambda_B\nabla_{\hat{x}_B}{\rm log}P(\hat{x}^B_i)\cdot[[{\rm log}P(x^A_i)-{\rm log}P(\hat{x}^A_i)]]_A$ and sends to \textbf{\emph{A}};\\
      Decrypts $[[{\rm log}\,P(x^B_i)-{\rm log}\,P(\hat{x}^B_i)]]_B$ for each $i$, calculates gradients of each layer using backpropagation;\\
      Updates the parameters $\theta_{BA}$ using optimizer $Opt_B$;\\

      \STATE \textbf{\emph{A}} do:\\
      Decrypts $[[{\rm log}\,P(x^A_i)-{\rm log}\,P(\hat{x}^A_i)]]_A$ for each $i$, calculates gradients of each layer using backpropagation;\\
      Updates the parameters $\theta_{AB}$ using optimizer $Opt_A$;\\
  \UNTIL{models converge}
  \RETURN $\theta_{AB}$ and $\theta_{BA}$
  \end{algorithmic}
\end{algorithm}
In a word, compared with conventional dual learning, the privacy-preserving dual learning scheme introduces a secure gradient descent-based backpropagation approach to update models by utilizing the Laplace mechanism and the additively homomorphic encryption. The former is to avoid the privacy disclosure of raw data in the training phase of dual models, while the latter is to prevent other participants from deriving the probability distribution of data that is implicit in the gradients, which further reduces the risk of data privacy leakage. The detailed privacy analysis is given in Section~\ref{ana}.

\subsubsection{Secure Multi-Party Learning with Collaborator}
\label{sec:multi}

The dual supplement approach can be regarded as a general solution for insufficient overlapping samples in vertical scenario, thus most conventional feature-based vertical federated learning methods \cite{yang2019federated} become available and can be transferred directly. In this way, the raw data of party $A$ and $B$ are kept locally, and the data interaction in the training process is secure under the definition. That is, any participant cannot acquire information from the other party beyond what is revealed by the input and output. Besides, dual supplement approach relaxes restrictions on vertical learning methods, as participants cannot derive further information other than output features of the affine transformation layer preserved differential privacy.

A simple but effective approach is introduced to instantiate the scheme. To be specific, following the idea of private layers in dual models, each participant holds its own perturbed affine transformation layer, which serves as the input layer. As depicted in Fig.~\ref{fig:thirdparty}, in the training stage of the joint learning task, the collaborator receives intermediate results, i.e., the weighted outputs of affine transformation layers deployed in participants, and calculates loss in the normal way to update the central layers. Afterwards, it sends the corresponding errors to each participant, respectively, to assist in updating the local layers.

Take the Multi-Layer Perception (MLP) \cite{Tang2016ExtremeLM} as an example, in the forward propagation, the weighted input $\{z_j\}_A=\sum_{k\in m_A} x_kw_{jk}$ and $\{z_j\}_B=\sum_{k\in m_B} x_kw_{jk}$ for the $j^{\rm th}$ neuron in the first hidden layer are sent to $C$, and $C$ calculates the sum $z_j=\{z_j\}_A+\{z_j\}_B$ and decrypts it for subsequent calculations. In the backward propagation, the partial derivatives of weights $\frac{\partial\mathcal{L}}{\partial w_{jk}^l}$ and biases $\frac{\partial\mathcal{L}}{\partial b_{j}^l}$ are with respect to an intermediate quantity $\delta_j^l$, which is the error of the $j^{\rm th}$ neuron in the $l^{\rm th}$ layer. The intermediate quantity is associated with errors and weights in the $(l+1)^{\rm th}$ layer:
\begin{equation}
\label{equ:back}
\delta^l_j=\sigma'(z_j^l)\sum\limits_{k=1}^m(\delta_k^{l+1}w_{kj}^{l+1}),
\end{equation}
where $w_{jk}^l$ denotes the weight for the connection between the $k^{\rm th}$ neuron in the $(l-1)^{\rm th}$ layer and the $j^{\rm th}$ neuron in the $l^{\rm th}$ layer, $a_k^l$ is the output of the $k^{\rm th}$ neuron in the $l^{\rm th}$ layer, and $\phi'(z_j^l)$ is the derivative of the activation function $\phi(\cdot)$ with the weighted input $z_j^l$. Without affecting the computations inside the central model, the collaborator $C$ sends errors $\delta_j$ for each neuron in the first hidden layer to party $A$ and $B$ for calculating gradients from the local affine transformation layer to the first hidden layer in $C$. The process is secure as even though the perturbed input features $x_k$ for each neuron in the affine transformation layer, which is preserved by differential privacy, are inferred by the other party, it cannot derive the weights $w_{jk}$ from what is obtains $\{z_j\}=\sum_{k} x_kw_{jk}$ based on the inability of solving $n$ equations with more than $n$ unknowns \cite{du2004privacy} \cite{vaidya2002privacy}.

\subsubsection{Discussion}
\label{ana}
The privacy and security properties include data privacy in dual learning and computational security in multi-party learning. Since the co-occurrence sample IDs are owned by both parties, the dual learning may raise two privacy issues: 1) a precise inference may lead to direct privacy disclosure, and 2) the dual loss contains information about data distribution. For the former, we inject identical Laplace noise into all features as a preprocessing step and build an affine transformation layer, thus each feature is preserved by the feature-oriented differential privacy with Laplace mechanism. At inference time, each party can only derive the differentially perturbed output of the affine transformation layer $\mathcal{F}(\mathcal{D})$ instead of the raw private features $\mathcal{D}$. In this way, participants cannot distinguish two neighboring datasets differing at most one feature in the other party, and further, whether there exist specific features in the dataset. For the latter, the probability distribution in the transmitted gradients (see Eq.~\ref{equ:longeq}) is encrypted by homomorphic encryption, which further makes it futile to derive the private information about the other's data, i.e., specific features, and the dual loss can be calculated and dual models are trained in a privacy-preserving manner. Extensive experiments are conducted as well to investigate how the differential privacy protects data privacy in the experimental section.

For the computational security defined in multi-party learning, the training and testing protocol of the central model does not reveal any information. What each participant can derive from the other party to the best of its ability are features that have been predicted in the dual learning, which are preserved by differential privacy, since all operations are based on the output of the differentially private affine transformation layer and have no direct access to the raw data. During training, party $A$ learns its own gradients of its affine transformation layer at each step, whereas it is not enough for $A$ to learn more information from $B$ based on the inability of solving $n$ equations with more than $n$ unknowns. Specifically, the number of supplementary overlapping samples $N_B$ is much greater than the number of features. In other words, there exists infinite number of inputs from $B$ to provide the same gradients to $A$. Similarly, $B$ can learn no information about $A$. Both parties are able to backpropagate and update the network parameters based on these gradients, hence they can only obtain the model parameters associated with their own features. As a result, each party remains oblivious to raw features of the other party at the end of the training process. 

In summary, we provide data privacy, computational security and performance gains in the proposed MPDL framework. Participant $A$, $B$ and collaborator $C$ hold the local model $\mathcal{M}_A$, $\mathcal{M}_B$ and $\mathcal{M}_F$ respectively, and only single-layer intermediate quantities are exchanged for the model updating, so that raw data and model structure are never exposed. Dual models provide high-quality supplementary data for the central model in the collaborator, which overcomes the challenge of insufficient overlapping samples and achieves superior performance over non-distributed self-learning models.

\section{Experiments}
\label{para:4}
In this section, several experiments are designed to validate the proposed MPDL approach. Specifically, we conduct experiments on image datasets to study the effectiveness and scalability of MPDL with various central models, such as MLP and CNN. We also employ widely used real-world datasets on banking and healthcare to evaluate the performance of our algorithm. Moreover, we verify the effectiveness of the proposed approach in protecting privacy on graphs.

\subsection{Dataset}
The MNIST dataset \cite{deng2012mnist} has 70000 handwritten digits from number 0 to 9. Based on the original black and white (bi-level) images from NIST, the resulting images in MNIST contain grey levels due to the anti-aliasing techniques and are centered in a $28\times 28$ image by computing the center of all pixels, and the image is translated to position this point at the center of the $28\times 28$ field.

The CIFAR-10 dataset \cite{alex:2009} consists of 60000 color images in 10 classes with 6000 images per class. The classes are completely mutually exclusive, and there is no overlap among them. Each image is $32\times 32$ with three channels.

The Bank Marketing dataset \cite{moro2014data} is related to direct marketing campaigns of a Portuguese banking institution, and the target is to predict whether the client will subscribe to a term deposit. The bank dataset has 45211 samples and 20 features, including age, job, marital, education, loan, etc.

The Breast Cancer Wisconsin (Diagnostic) dataset \cite{Dua:2019} contains 569 instances and 32 different features, which are computed from digitized images of a fine needle aspirate of a breast mass, and they describe characteristics of the cell nuclei present in images, such as radius, texture and concave points. There are 357 benign and 212 malignant in the dataset.

The Google+ dataset \cite{leskovec2012learning} is an ego-network of Google+ users, which it contains 1206 nodes and 66918 links, and nodes in the network represent the user's friends. Each node is described by a 940-dimensional vector constructed from tree-structured user profiles, and the user's gender is treated as the class label.

The Hamilton dataset \cite{leskovec2012learning} is a collection of US university Facebook networks, which consists of 2118 nodes and 87486 links. Each node is described by a 144-dimensional vector, representing anonymous personal information of them. The student status flag is used as the class label, including a total of five unbalanced categories.

\subsection{Experimental Settings}

The datasets are split both in the feature space and the sample space to simulate a two-party distributed learning problem, and the two parts are stored separately on two local servers. Another local server, which is assumed to be the Trusted Execution Environment, serves as the participant $C$. For the first four datasets, we assign all the labels to party $B$, and take 10\% of the data as test samples that contain all features. For the rest samples $\mathcal{D}=\{(x_i^A,x_i^B,y_i^B)\}_{i\in N}$, a hyperparameter $\gamma$ is introduced to control the co-occurrence probability, and we have $\mathcal{D}_C=\{(x_i^A,x_i^B,y_i^B)\}_{i\in N\cdot[0,\,\gamma]}$ for dual learning, $\mathcal{D}_B=\{(x_i^B,y_i^B)\}_{i\in N\cdot(\gamma,\,0.5+\gamma/2]}$ for further central training, and $\mathcal{D}_A=\{(x_i^A)\}_{i\in N\cdot(0.5+\gamma/2,\,1])}$ for validating the central model. The number of folds $K$ is set to 5. Note that all samples are shuffled and there is no guarantee that a balanced ratio among different classes is maintained, so that the data distribution is similar to industrial scenarios.

Other key impacting factors are threshold $T$ and max iterations $m$ that affect the training effect of dual models. The threshold $T$ represents the desired improvement, and once it is achieved on the validation set, the models are considered to converge and the training is terminated. Due to the double-trigger mechanism that controls the terminal condition, the range of $T$ is flexible and a value greater than 0.1 is usually taken. Even though the threshold could be unreachable, as the training will also stop at the max iteration m. For the MNIST, Breast Cancer, Google+ and Hamilton datasets, we set we set $T = 0.15$ and $m = 2$, and we have $T = 0.1$ and $m = 4$ 4 for the CIFAR-10 and Bank Marketing datasets as their data distribution is more complicated. Moreover, the privacy budget $\epsilon$ in the accuracy evaluation part is fixed at 0.5 for the for the simple datasets and 2 for the complicated ones. It controls the amount of noise injected into features, and a smaller privacy budget value enforces a stronger privacy guarantee. For more complicated dataset, the privacy budget consumption is relatively larger. Extensive privacy evaluation experiments are conducted to investigate how it affects the performance in Section~\ref{peva}.

Dual models are MLP with opposite input layer and output layer, which contain one hidden layer of $(N_{input}+N_{output})/2$ neurons. They use ReLU as the activation function and the mean-square error (MSE) as the loss function $\ell_{align}$. The Lagrange parameters $\lambda_A$ and $\lambda_B$ are set 0.01 by trial and error. Central MLP model has a similar structure as dual models, while the last layer is a multi-class softmax output layer, and the cross entropy error function is adopted. For MNIST dataset, CNN model has two $5\times 5$ kernel filters (the first with 16 channels and the second with 32 channels) followed by two fully connected layers and a 10 class softmax output layer. For CIFAR-10 dataset, CNN model has three $3\times 3$ kernel filters, which have 16, 32, 32 channels respectively and followed by a $2\times 2$ max pooling layer, and it also has two fully connected layers as MNIST. The minibatch SGD algorithm with the learning rate of 0.1 is employed to optimize both MLP and CNN models, and the training epochs are 10 for dual learning and 20 for central training.

Since the privacy-preserving feature-based learning is lossless compared with a jointly built model without privacy constraints \cite{yang2019federated}, we adopt it as a baseline while dealing with the condition that there is no missing feature, i.e., the test set, and the non-distributed central model for it (named ${\rm joint}_T$) is the same as ours (${\rm dual}_T$). Note that the only difference between them is that ${\rm joint}_T$ is merely trained on co-occurrence samples $\mathcal{D}_C$ while ${\rm dual}_T$ is further trained on generated supplementary data $\mathcal{D}_B$. For the samples $\mathcal{D}_A$ whose features are only held by one party, the federated transfer learning model (${\rm FTL}_A$) is used as a baseline, and ours is recorded as ${\rm MPDL}_A$. For ${\rm FTL}_A$, two stacked auto-encoder layers are trained for each party separately, and the central model is MLP of the same layers as ${\rm MPDL}_A$.

\begin{figure}[!tb]
  \centering
  \includegraphics[width=0.6\linewidth]{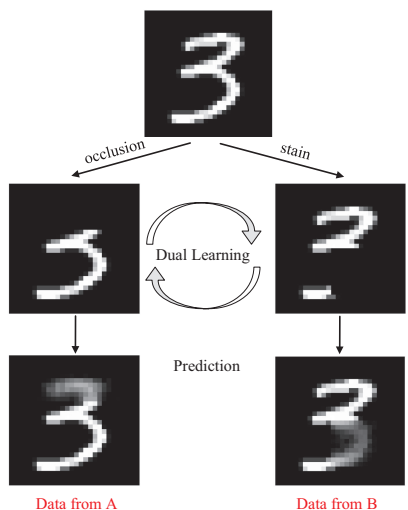}
\caption{The process of damaged images inpainting by dual models.}
\label{fig:mnist}
\end{figure}

In the experiments, we repeated the trial for ten times and the average classification accuracy with different co-occurrence probability $\gamma$ on the first four datasets is shown in Tables~\ref{tab:mnist}-\ref{tab:cancer}. The proposed ${\rm MPDL}_A$ can retain the original structure of images after mapping $\{(x_i^A)\}_{i\in N_A}$ to $\{(x_i^A,x_i^B)\}_{i\in N_A}$, see Fig.~\ref{fig:mnist}, whereas ${\rm FTL}_A$ learns a low-dimensional vector of images, therefore the CNN model cannot be carried out on ${\rm FTL}_A$.

\subsection{Application to Images Processing}

\begin{table*}[!tp]
\footnotesize
\renewcommand{\arraystretch}{1.3}
  \centering
  \caption{Comparison with baselines on MNIST.}
  \setlength{\tabcolsep}{4mm}{
    \begin{tabular}{p{1mm}ccccccc}
    \cline{2-8}
      &\multirow{2}{*}{\bf{Methods}}&\multicolumn{6}{c}{\bf{co-occurrence probability $\gamma$}}\\
      \cline{3-8}
      &&0.05&0.1&0.2&0.4&0.6&0.8\\
      \cline{2-8}
      \multirow{4}{*}{\rotatebox{90}{MLP}}&${\rm joint}_T$&66.43$\pm$2.40&67.45$\pm$2.31&67.76$\pm$2.17&76.92$\pm$1.98&85.15$\pm$1.54&88.12$\pm$1.45\\
      &${\rm dual}_T$&74.85$\pm$2.06&75.98$\pm$2.30&78.55$\pm$2.60&83.98$\pm$2.25&85.74$\pm$2.39&87.36$\pm$2.15\\
      &${\rm FTL}_A$&46.29$\pm$4.62&52.56$\pm$5.74&59.62$\pm$5.70&64.98$\pm$4.48&72.79$\pm$4.32&83.35$\pm$4.14\\
      &${\rm MPDL}_A$&74.11$\pm$3.23&74.54$\pm$2.61&75.03$\pm$2.79&80.17$\pm$3.49&83.01$\pm$3.37&84.26$\pm$3.63\\
        \cline{2-8}
        \multirow{3}{*}{\rotatebox{90}{CNN}}&${\rm joint}_T$&82.82$\pm$2.83&84.66$\pm$3.10&85.91$\pm$2.65&91.68$\pm$2.10&92.16$\pm$2.53&94.96$\pm$1.62\\
      &${\rm dual}_T$&91.15$\pm$3.72&92.53$\pm$2.64&94.49$\pm$2.98&95.52$\pm$2.99&96.89$\pm$2.17&97.04$\pm$1.47\\
      &${\rm MPDL}_A$&86.22$\pm$2.87&88.95$\pm$3.43&90.03$\pm$2.89&91.25$\pm$2.78&91.73$\pm$2.83&93.23$\pm$2.13\\
      \cline{2-8}
    \end{tabular}}
\label{tab:mnist}
\end{table*}

\begin{table*}[tbp]
\footnotesize
\renewcommand{\arraystretch}{1.3}
  \centering
  \caption{Comparison with baselines on CIFAR-10.}
  \setlength{\tabcolsep}{4mm}{
    \begin{tabular}{p{1mm}ccccccc}
    \cline{2-8}
      &\multirow{2}{*}{\bf{Methods}}&\multicolumn{6}{c}{\bf{co-occurrence probability $\gamma$}}\\
      \cline{3-8}
      &&0.05&0.1&0.2&0.4&0.6&0.8\\
      \cline{2-8}
      \multirow{4}{*}{\rotatebox{90}{MLP}}&${\rm joint}_T$&26.07$\pm$2.26&29.01$\pm$2.39&34.96$\pm$1.82&35.14$\pm$1.53&37.07$\pm$2.26&44.88$\pm$1.97\\
      &${\rm dual}_T$&30.76$\pm$2.56&36.33$\pm$2.55&41.25$\pm$2.96&41.25$\pm$2.53&42.21$\pm$2.87&46.61$\pm$2.50\\
      &${\rm FTL}_A$&16.10$\pm$1.46&22.62$\pm$3.72&32.78$\pm$4.12&33.42$\pm$2.69&35.10$\pm$3.44&38.86$\pm$2.16\\
      &${\rm MPDL}_A$&26.83$\pm$2.63&30.10$\pm$3.79&35.36$\pm$2.75&38.55$\pm$3.47&40.53$\pm$2.94&43.52$\pm$2.70\\
        \cline{2-8}
        \multirow{3}{*}{\rotatebox{90}{CNN}}&${\rm joint}_T$&44.07$\pm$2.51&47.76$\pm$3.10&54.19$\pm$3.75&60.88$\pm$2.01&64.73$\pm$2.76&65.45$\pm$2.51\\
      &${\rm dual}_T$&57.09$\pm$2.60&57.74$\pm$3.40&58.91$\pm$3.23&60.37$\pm$3.13&63.92$\pm$3.71&68.66$\pm$2.92\\
      &${\rm MPDL}_A$&50.32$\pm$3.51&52.62$\pm$3.09&55.74$\pm$3.71&57.66$\pm$3.84&59.26$\pm$3.13&62.84$\pm$3.26\\
      \cline{2-8}
    \end{tabular}}
\label{tab:cifar}
\end{table*}

\begin{figure*}[!t]
  \centering
  \subfigure{
    \includegraphics[width=0.3\linewidth]{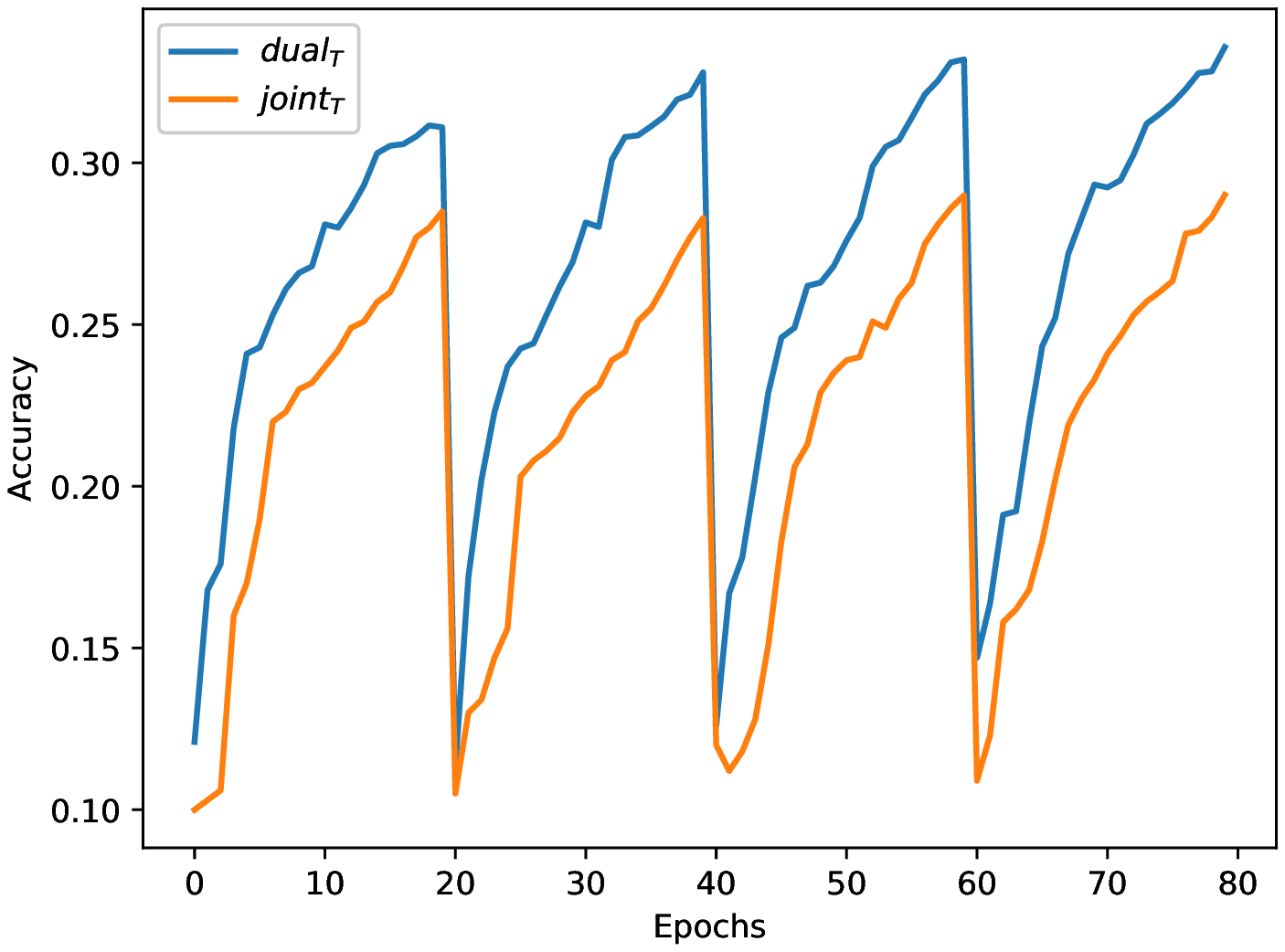}}
    \qquad\qquad
  \subfigure{
    \includegraphics[width=0.3\linewidth]{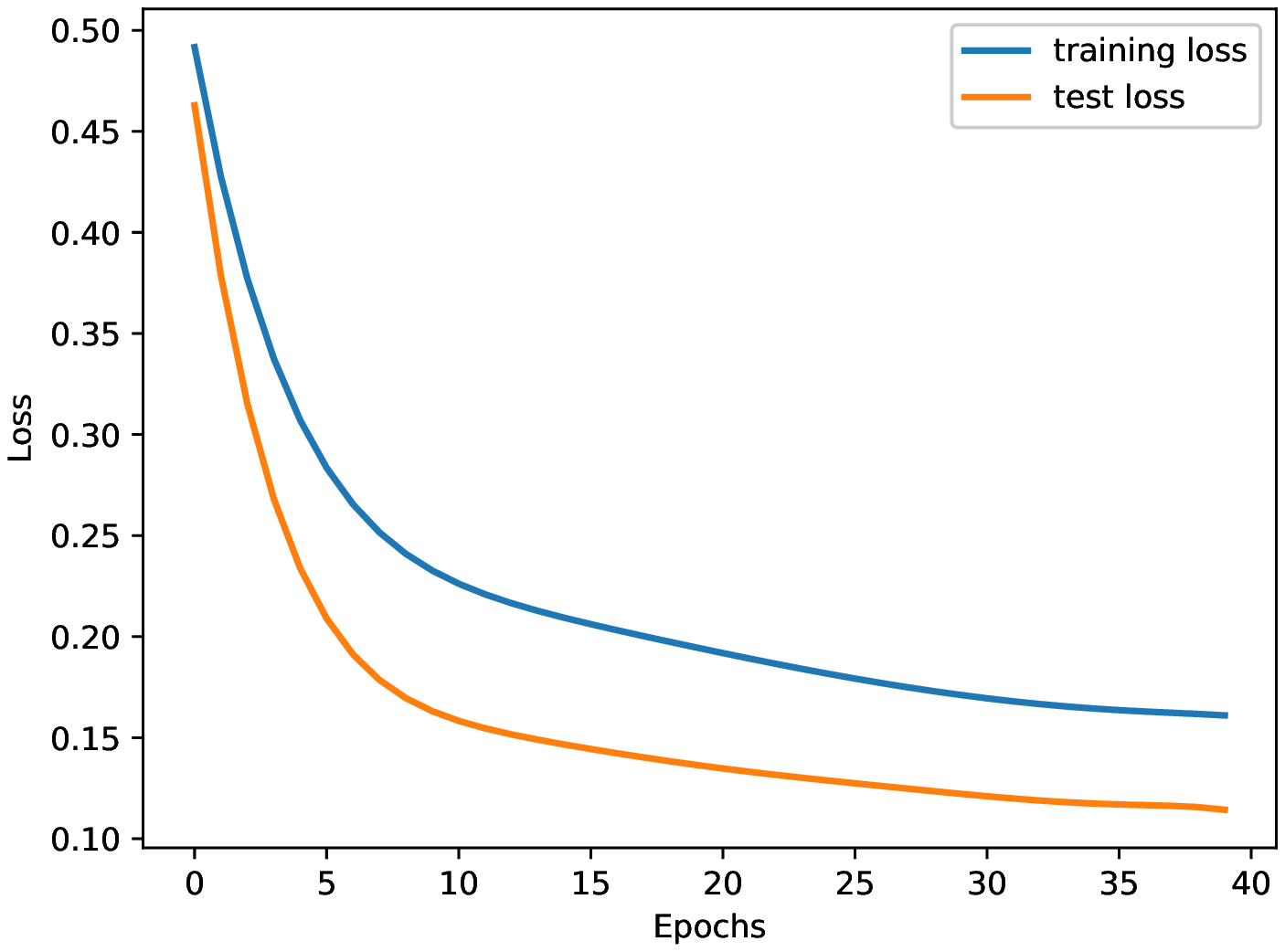}}
  \caption{Performance of the central model and dual models on CIFAR-10 during training phase: the accuracy of the ${\rm joint}_T$ and ${\rm dual}_T$ on the test set (Left); training loss and test loss of duals models (Right).}
\label{fig:line}
\end{figure*}
For the MNIST and CIFAR-10 datasets, we simulate the scene that images in both parties are damaged or missing to varying degrees (e.g., the masked regions are due to object occlusion, broken equipment, stain pollution or other problems during the process of shooting or storage). For the MNIST dataset, we assume images in party $A$ only holds the bottom half $18\times 28$ pixels of images due to external object occlusion for the camera, and images in party $B$ are stained in the lower right corner of $14\times 16$ pixels, see Fig.~\ref{fig:mnist}. For the CIFAR-10 dataset, the corresponding region sizes are $20\times 32$ and $16\times 18$, respectively. Meanwhile, we assume that both participants have prior knowledge of the other party's damaged location through communication. As the structure of original images is preserved through image inpainting, we verify the effectiveness of CNN on the image datasets.

Table~\ref{tab:mnist} shows the classification accuracy of different algorithms on MNIST. The proposed scheme $dual_T$ performs significantly better than $joint_T$ when the co-occurrence probability $\gamma$ is small. Moreover, with the increase of $\gamma$, the performance difference between the two models is gradually narrowing. To be specific, when $\gamma=0.05$, there are 5\% of the data in the dataset overlaps in party $A$ and $B$, and $joint_T$ can merely be trained on this part of data. However, dual models are able to expand 47.5\% of the samples that are labeled and only exist in party $B$ to co-occurrence pairs, and the central model can leverage 52.5\% of the training samples. Therefore, the proposed framework achieves a higher accuracy than a non-private scheme, though there may exist inference error and performance loss due to differential privacy. Similarly, the performance of MPDL is more robust compared with FTL on $\mathcal{D}_A$, which indicates that the MPDL model is capable of inferring and classifying the missing data with very few overlapping pairs. When CNN serves as the central model, the three algorithms achieve higher classification accuracy. Due to more training data for the CNN and the improved generalization performance of the generated samples, ${\rm MPDL}_A$ outperforms ${\rm joint}_T$ even with merely half of the features when training data is less than 80\%. 

\begin{table*}[!tbp]
\footnotesize
\renewcommand{\arraystretch}{1.3}
  \centering
  \caption{Comparison with baselines on Bank Marketing.}
  \setlength{\tabcolsep}{4mm}{
    \begin{tabular}{p{1mm}ccccccc}
    \cline{2-8}
      &\multirow{2}{*}{\bf{Methods}}&\multicolumn{6}{c}{\bf{co-occurrence probability $\gamma$}}\\
      \cline{3-8}
      &&0.05&0.1&0.2&0.4&0.6&0.8\\
      \cline{2-8}
      \multirow{4}{*}{\rotatebox{90}{MLP}}&${\rm joint}_T$&79.23$\pm$1.28&80.36$\pm$1.34&81.77$\pm$2.04&87.11$\pm$7.54&93.76$\pm$4.02&96.63$\pm$2.09\\
      &${\rm dual}_T$&88.03$\pm$5.53&89.61$\pm$3.27&92.76$\pm$3.53&94.44$\pm$3.11&94.86$\pm$3.11&95.95$\pm$2.96\\
      &${\rm FTL}_A$&80.48$\pm$1.37&79.14$\pm$1.10&79.76$\pm$1.73&80.05$\pm$1.35&79.82$\pm$1.38&79.63$\pm$0.86\\
      &${\rm MPDL}_A$&82.86$\pm$4.54&84.46$\pm$3.92&86.01$\pm$3.69&88.35$\pm$4.10&91.52$\pm$3.85&92.28$\pm$2.95\\
        \cline{2-8}
    \end{tabular}}
\label{tab:bank}
\end{table*}

\begin{table*}[!thp]
\footnotesize
\renewcommand{\arraystretch}{1.5}
  \centering
  \caption{Comparison with baselines on Breast Cancer.}
  \setlength{\tabcolsep}{4mm}{
    \begin{tabular}{p{1mm}ccccccc}
    \cline{2-8}
      &\multirow{2}{*}{\bf{Methods}}&\multicolumn{6}{c}{\bf{co-occurrence probability $\gamma$}}\\
      \cline{3-8}
      &&0.05&0.1&0.2&0.4&0.6&0.8\\
      \cline{2-8}
      \multirow{4}{*}{\rotatebox{90}{MLP}}&${\rm joint}_T$&78.26$\pm$3.18&82.61$\pm$3.13&84.35$\pm$3.90&89.53$\pm$3.61&91.30$\pm$2.51&93.04$\pm$2.42\\
      &${\rm dual}_T$&85.81$\pm$3.04&86.64$\pm$2.11&89.41$\pm$2.71&90.11$\pm$2.14&91.38$\pm$2.80&94.02$\pm$1.54\\
      &${\rm FTL}_A$&66.99$\pm$6.03&65.74$\pm$7.04&68.47$\pm$4.35&81.88$\pm$4.22&85.91$\pm$3.12&87.23$\pm$3.23\\
      &${\rm MPDL}_A$&81.02$\pm$4.69&84.07$\pm$3.59&85.83$\pm$3.42&88.34$\pm$3.18&90.86$\pm$2.90&91.46$\pm$2.53\\
        \cline{2-8}
    \end{tabular}}
\label{tab:cancer}
\end{table*}

Table~\ref{tab:cifar} represents the classification accuracy on CIFAR-10. Each image in CIFAR-10 contains 3072 pixels while there are only 784 pixels in MNIST, hence it is difficult for MLP model to identify images from different classes. Nevertheless, MPDL still outperforms the joint model and FTL model, especially with limited co-occurrence samples. 
When there are sufficient overlapping pairs ($\gamma=0.8$), ${\rm dual}_T$ that leverages all features achieves the highest average accuracy, and the performance of ${\rm MPDL}_A$ is slightly worse than ${\rm joint}_T$. The reason for the situation is that dual models are unable to accurately infer missing data using features from one single party, because same features in party $A$ may correspond to different features in party $B$. For instance, it could be hard to distinguish between a cat and a dog if its head in the image happens to be stained. With the help of the CNN model, MPDL outperforms FTL by more than 35\% with limited training data and relatively 25\% with more samples.

The classification accuracy of the two models with MLP is shown in Fig.~\ref{fig:line} when $\gamma$ is set 0.1, and the $x$ axis indicates the actual training epoch regardless of continuity, because ${\rm joint}_T$ and ${\rm dual}_T$ are retrained every 20 epochs. In each training session, ${\rm dual}_T$ converges faster with more training samples. Meanwhile, dual models are constantly trained without reinitialization, and the quality of inferred data is also improving. In the four iterations, the accuracy of ${\rm joint}_T$ is approximately the same, whereas the performance of ${\rm dual}_T$ improves as the quality of co-occurrence samples increases. Loss decays at the same rate on the training set and test set, and it demonstrates the generalization baility of dual models and the reliability of the inferred data. Training loss is larger than test loss due to the  probabilistic duality constraint $\lambda \ell_{dual}$, which is not included in the test loss.

\subsection{Application to Data Analysis}
Features of the Bank Marketing and Breast Cancer dataset are assigned to each party randomly and none of them overlaps between parties $A$ and $B$. Moreover, one-hot encoding is applied to categorical features in Bank Marketing, and the processed dataset contains 48 features. The trial is repeated many times to simulate various financial and medical institutions.

Bank Marketing dataset has more practical significance since users' payment data (housing and personal loan) and profile data (job and education) can be held by different institutions. Note that the attributes are randomly assigned to each party and then one-hot encoding is performed on them. There are 36193 positive samples and 4594 negative samples in the training set, and the original ratio between them is maintained in order to simulate real scenarios.
The ${\rm FTL}_A$ model shows worst performance on this dataset, as its classification accuracy is merely equivalent to random guessing. The ${\rm joint}_t$ model cannot learn the intrinsic discriminating information either with less than 20\% co-occurrence samples. Under extremely unbalanced data distribution, the MPDL model converges well and the performance improves as increasing the number of co-occurrence samples.

\begin{figure*}[!t]
  \centering
  \subfigure{
    \includegraphics[width=0.3\linewidth]{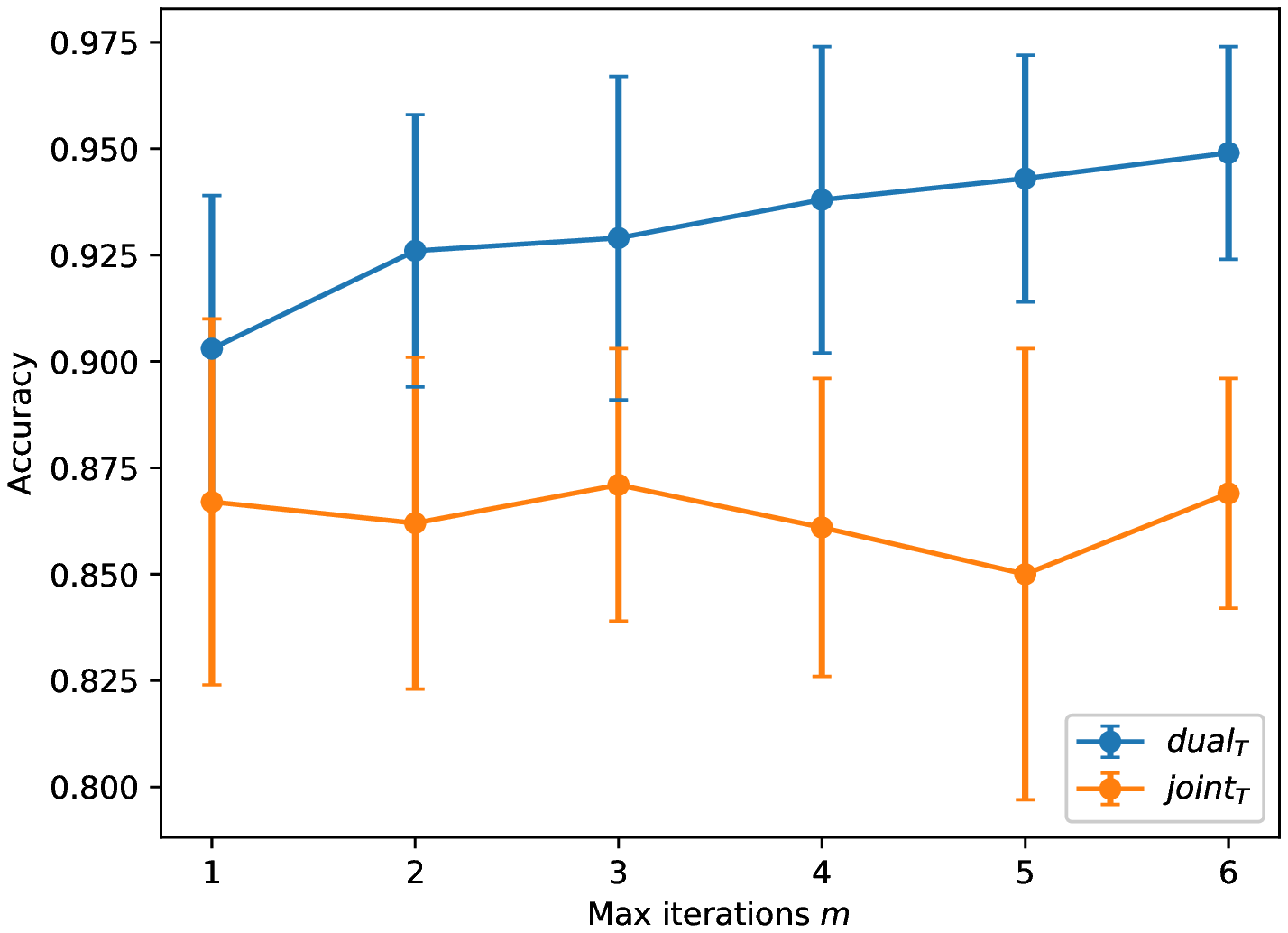}}
    \qquad\qquad
  \subfigure{
    \includegraphics[width=0.3\linewidth]{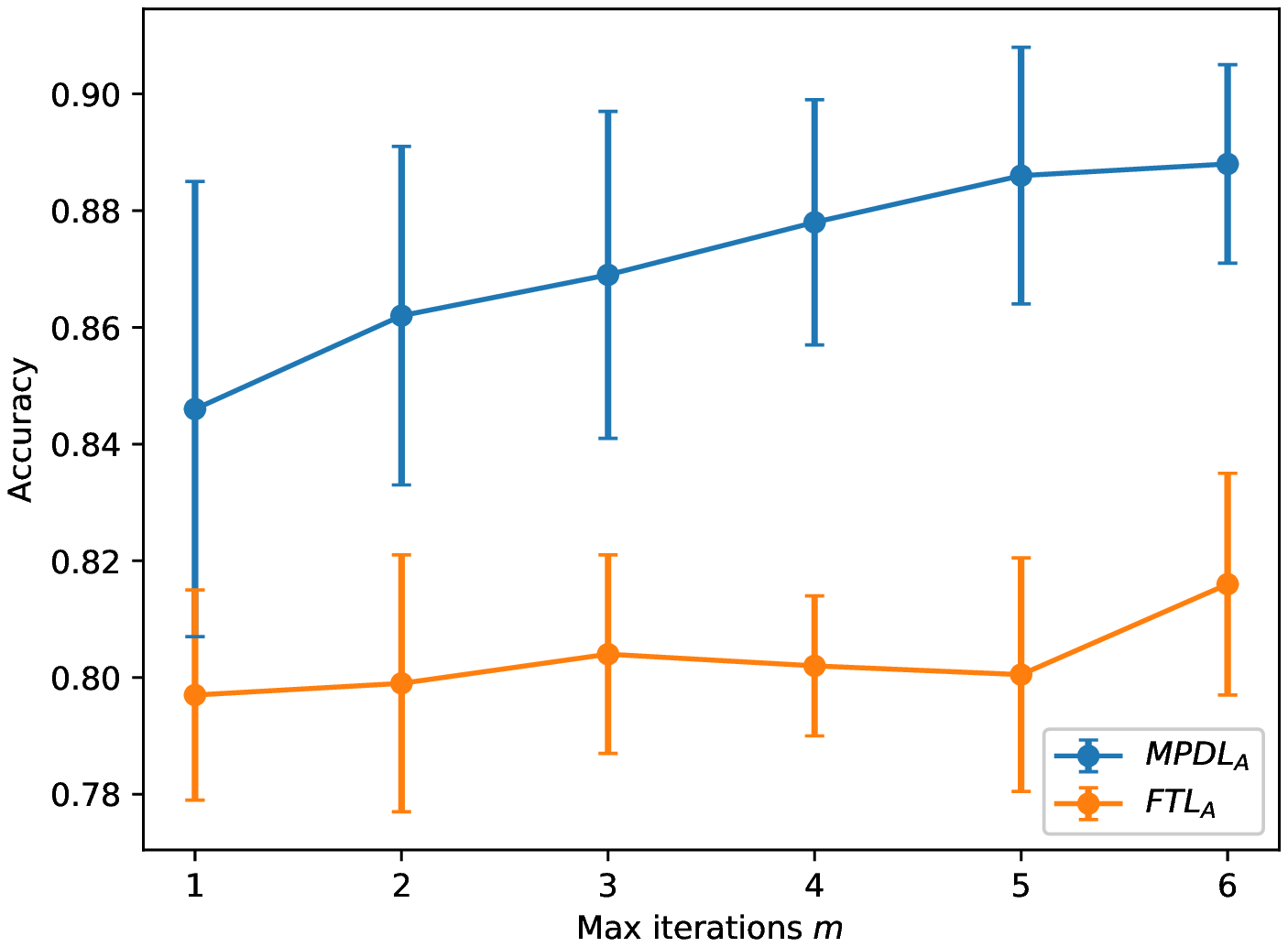}}
  \caption{Effect of max iterations $m$ on Bank Marketing: the accuracy of the ${\rm joint}_T$ and ${\rm dual}_T$ on the test set (Left); the accuracy of MPDL and FTL on samples $\mathcal{D}_A$ (Right).}
\label{fig:para}
\end{figure*}

\begin{figure*}[!tpb]
  \centering
  \subfigure{
    \includegraphics[width=0.28\linewidth]{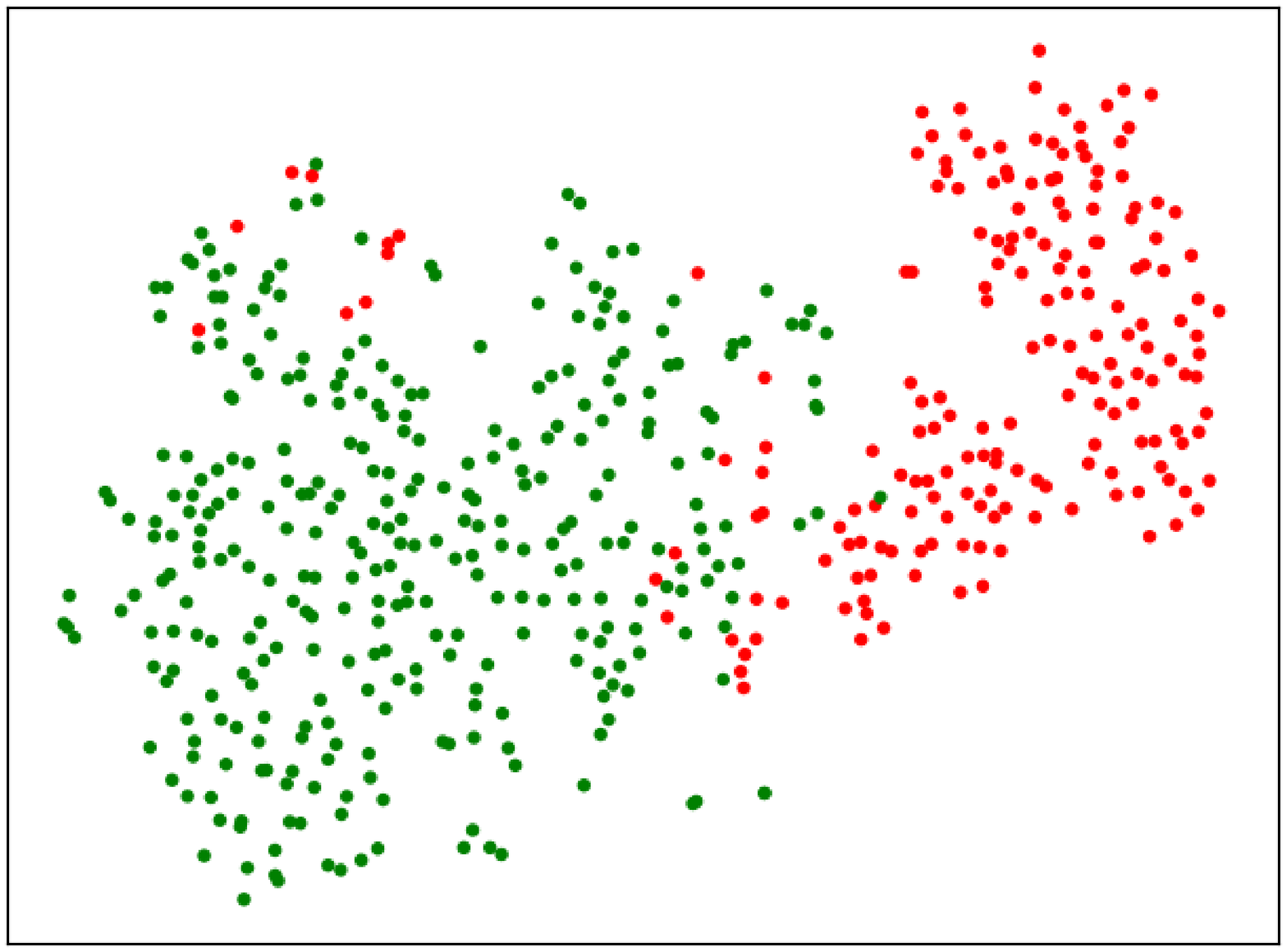}}
  \qquad
  \subfigure{
    \includegraphics[width=0.28\linewidth]{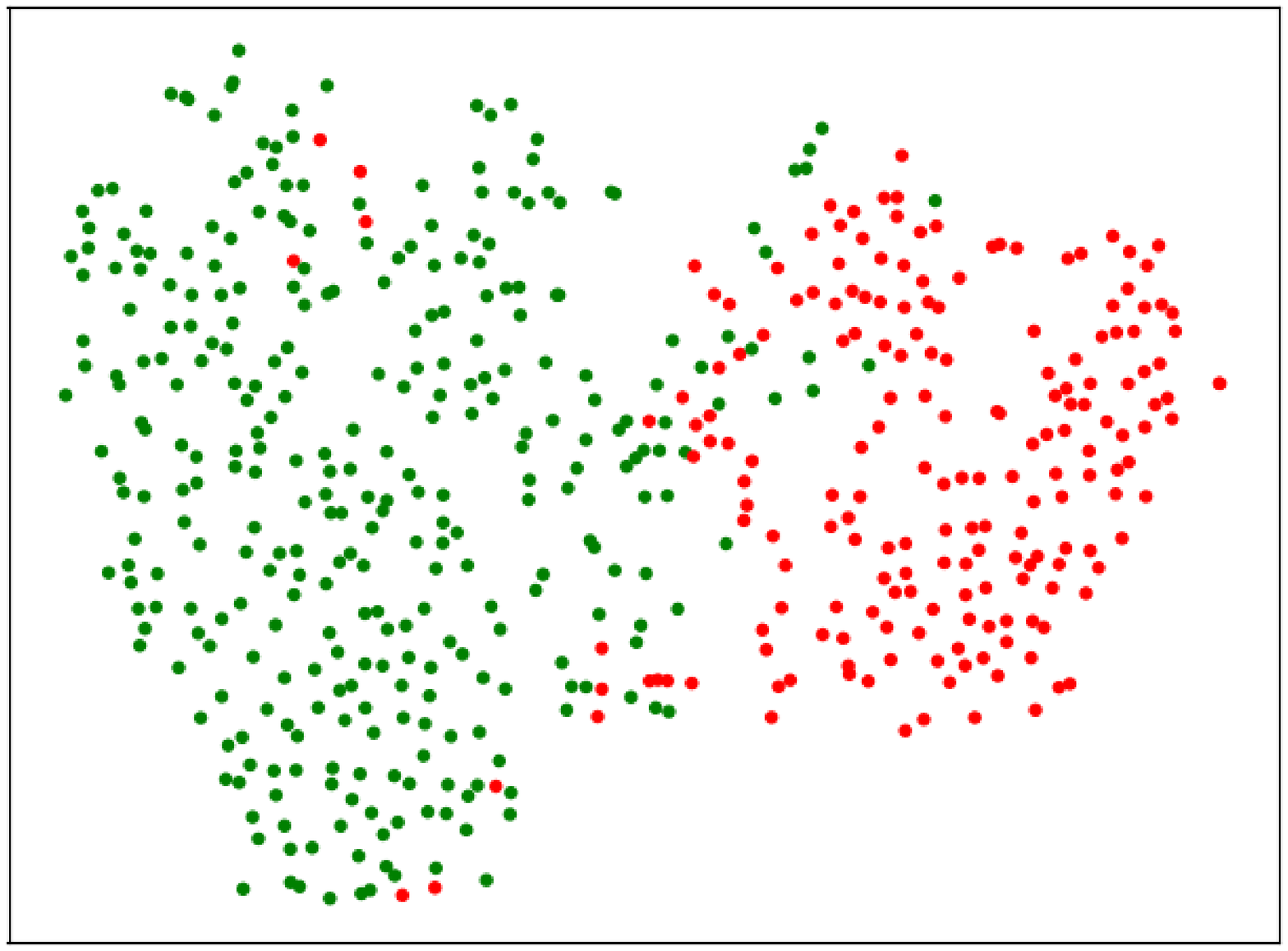}}
  \qquad
  \subfigure{
    \includegraphics[width=0.28\linewidth]{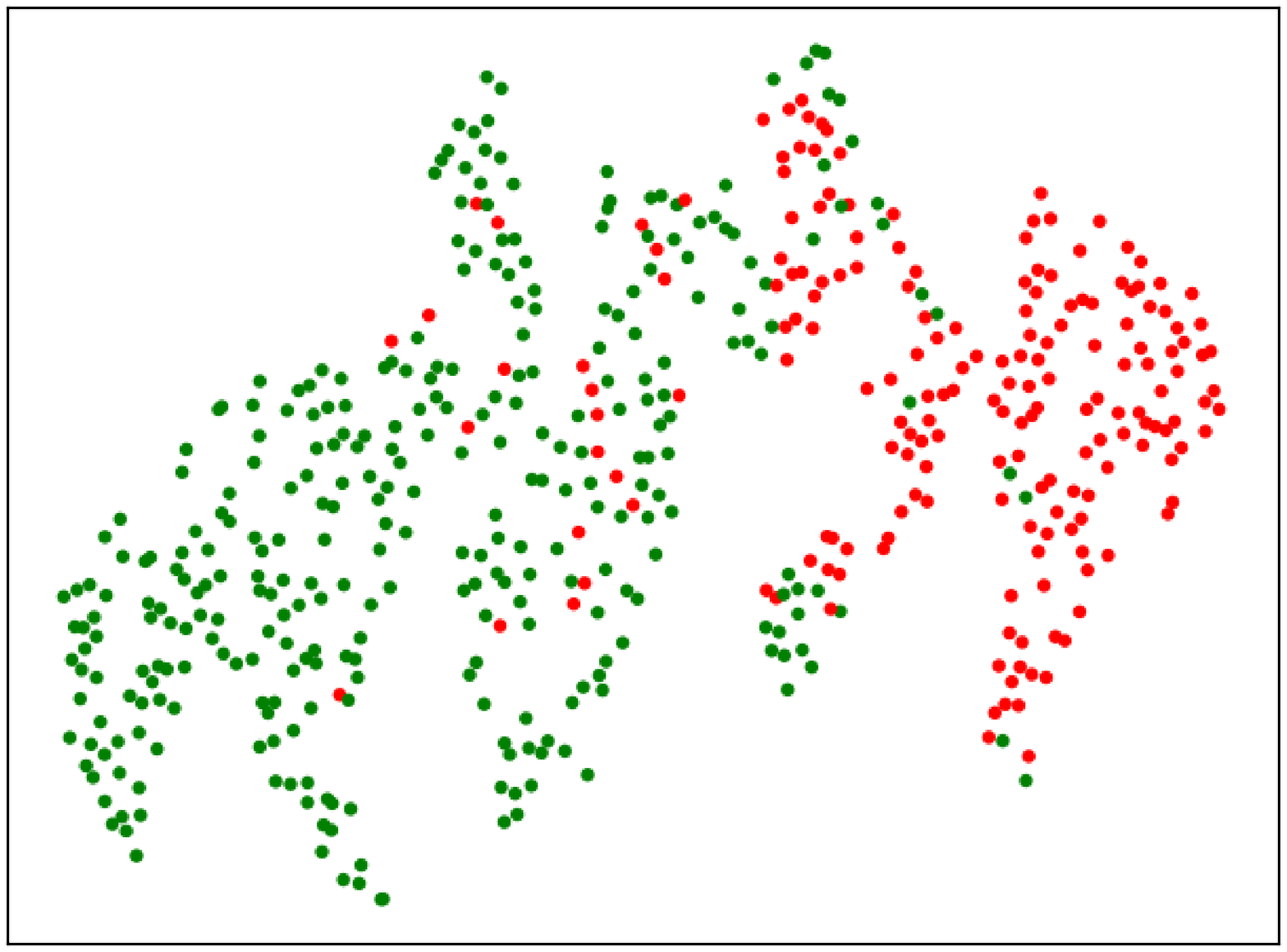}}
  \caption{Visualization of the Breast Cancer dataset with original data (Left), multi-party dual learning (Middle), and federated transfer learning (Right).}
\label{fig:vis}
\end{figure*}
We evaluate the performance of the proposed dual cross validation and measure the effect of max iterations $m$ on the Bank Marketing dataset. The threshold $T$ plays the same role as $m$, which serves as the basis to judge the convergence of dual models. The experiment is conducted using the MLP model and $\gamma=0.4$, and results are shown in Fig.~\ref{fig:para}. The dual models are constantly trained without reinitialization in iterations, therefore the performance of ${\rm dual}_t$ and ${\rm MPDL}_A$ improves as the number of iterations increase. Besides, for ${\rm dual}_t$, the reliability of inferred data only affects the training process, while for the ${\rm MPDL}_A$ model, it further influences the predict data and test process. The ${\rm joint}_t$ model is reinitialized so that $m$ has no effect on it, and ${\rm FTL}_A$ is limited by the bottleneck of the model. Since a new fold of data is selected for each iteration, the number of folds $K$ needs to be no less than the max iterations to improve the generalization performance of the model.

As mentioned before, medical data are very sensitive and private thus difficult to collect, and they exist in isolated hospitals and medical centers. The lack of data sources and the insufficiency of co-occurrence samples drag down the performance of central models. In Table~\ref{tab:cancer}, when there are less than 20\% overlapping pairs, the misjudgement rate of the FTL model on $\mathcal{D}_A$ is up to 35\%, which means one in third people get tested for cancer will miss the best opportunity for treatment or fall into an unnecessary panic and unease. Some significant features could be lost during the process of transferring information from the two parties into a common space, and it leads to unsatisfactory performance of the model. 

The visualization of the original data, inferred data of MPDL and the hidden representations of FTL are shown in Fig.~\ref{fig:vis}. These vectors are mapped into a two-dimensional space using t-SNE \cite{maaten2008visualizing}, which intuitively reveals the intrinsic structure of data. We set $\gamma=0.05$ and remaining 95\% of the data with half of the features is visualized for MPDL and FTL. The MPDL model is capable of forming a similar distribution as the original data, whereas the mapped representations of the FTL model significantly change the structure of data, and the information loss of knowledge transfer becomes the bottleneck of the federated transfer learning. The proposed MPDL model increases the classification accuracy by 15\% to 20\% with a small part of training data, and the performance is comparable to the joint model that predicts using all features.

\begin{figure*}[!t]
  \centering
  \subfigure{
    \includegraphics[width=0.35\linewidth]{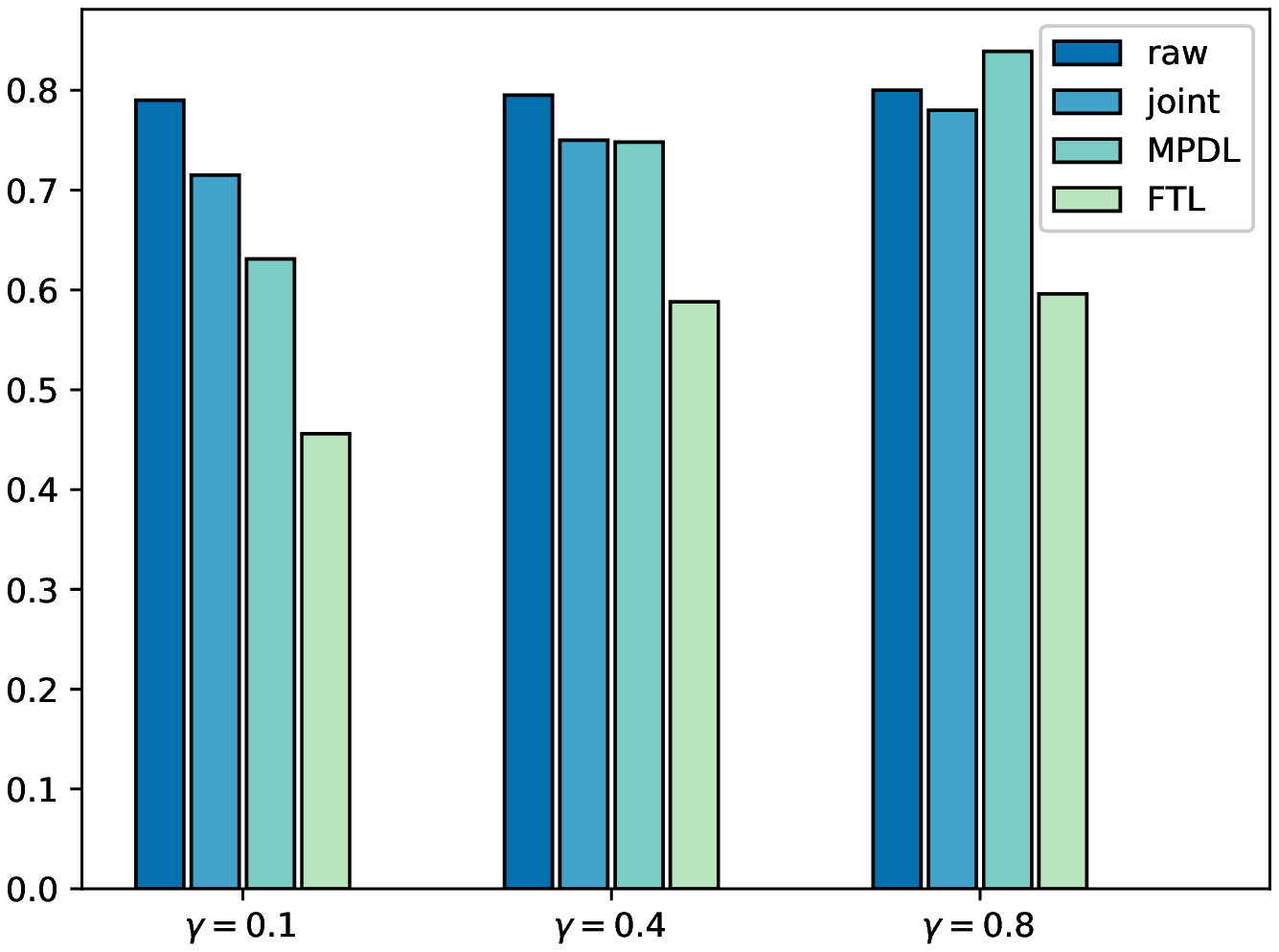}}
    \qquad\qquad
  \subfigure{
    \includegraphics[width=0.35\linewidth]{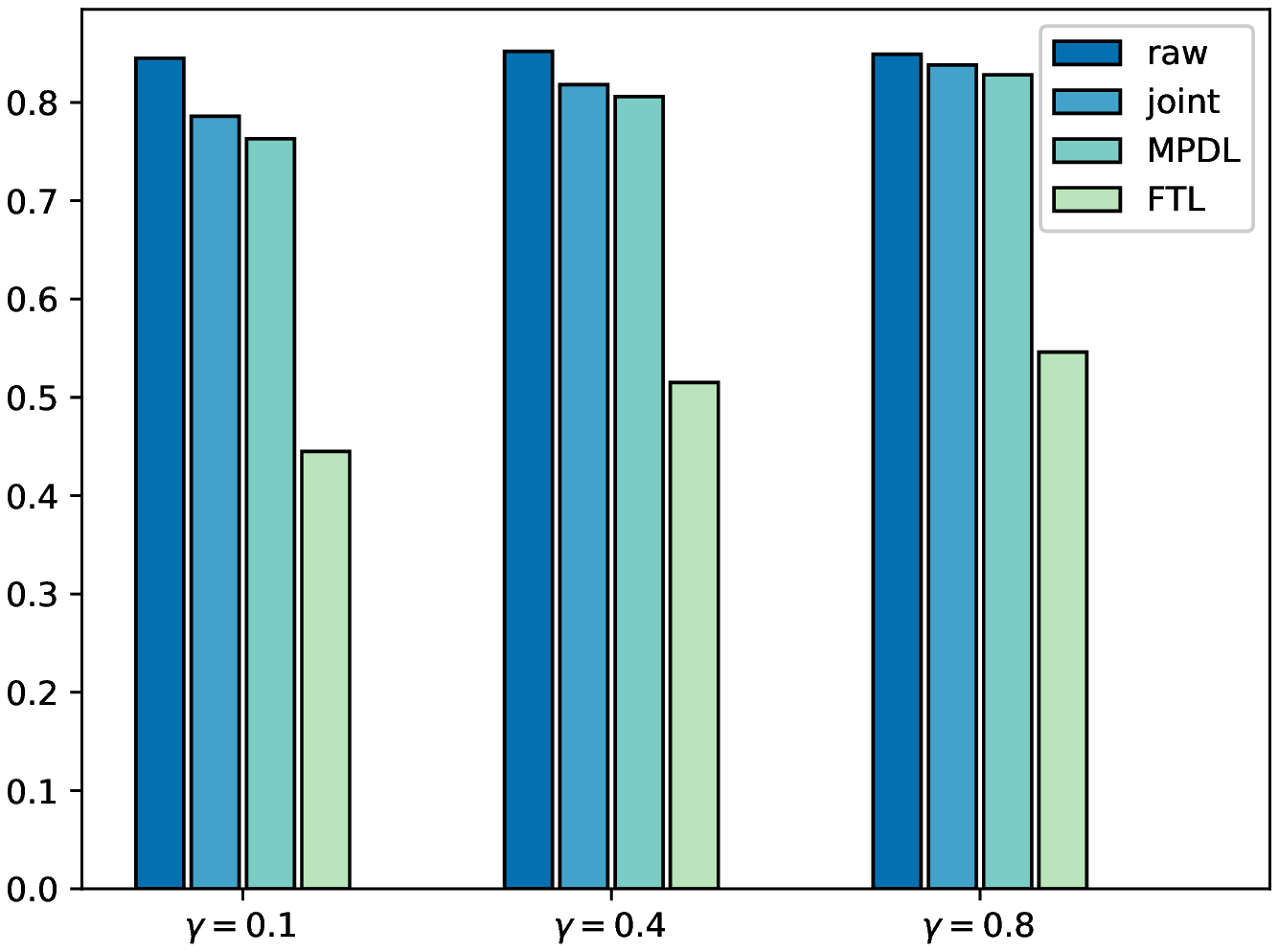}}
  \caption{AUC scores with different co-occurrence probabilities on Google+ (Left) and Hamilton (Right) for link prediction.}
\label{fig:link}
\end{figure*}

\subsection{Application to Graphs Processing}

\begin{figure}[!t]
  \centering
  \includegraphics[width=0.75\linewidth]{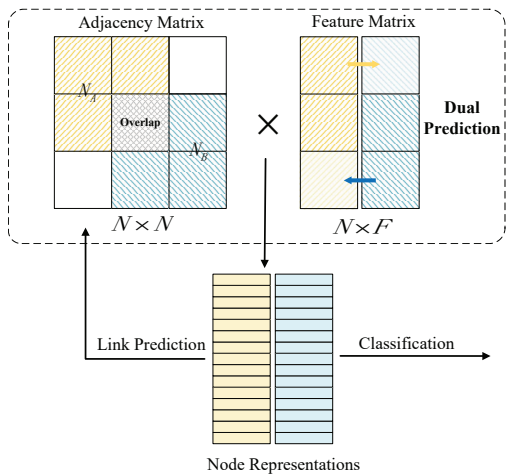}
\caption{The process of link prediction and node classification in graphs by dual models.}
\label{fig:graph}
\end{figure}
For the Google+ and Hamilton datasets, we simulate the scenario that party $A$ holds an information graph $G_A=(V_A,E_A,F_A)$, and party $B$ holds $G_B=(V_B,E_B,F_B)$, where $E \subseteq V\times V$ are links in the graph and $F$ represents personal information of nodes. In general, for a node $i$ in graph $G_A$, the personal information is denoted as $x_i^A$, and in graph $G_B$ it is $x_i^B$. Similar to the statement in Section~\ref{para:3}, we assume that there exists a limited set of co-occurrence nodes, thus there is an overlapping part between the two adjacency matrices, see Fig.~\ref{fig:graph}. Dual models are trained on features of co-occurrence pairs $\{(x_i^A,x_i^B)\}_{i\in N_C}$ and utilized for the feature matrix completion. Leveraging the inferred feature matrix, we could have node representations $M_{rep}$:
\begin{equation}
M_{rep}=M_{adj}\times M_{feat},
\end{equation}

\begin{figure*}[!t]
  \centering
  \includegraphics[width=0.9\linewidth]{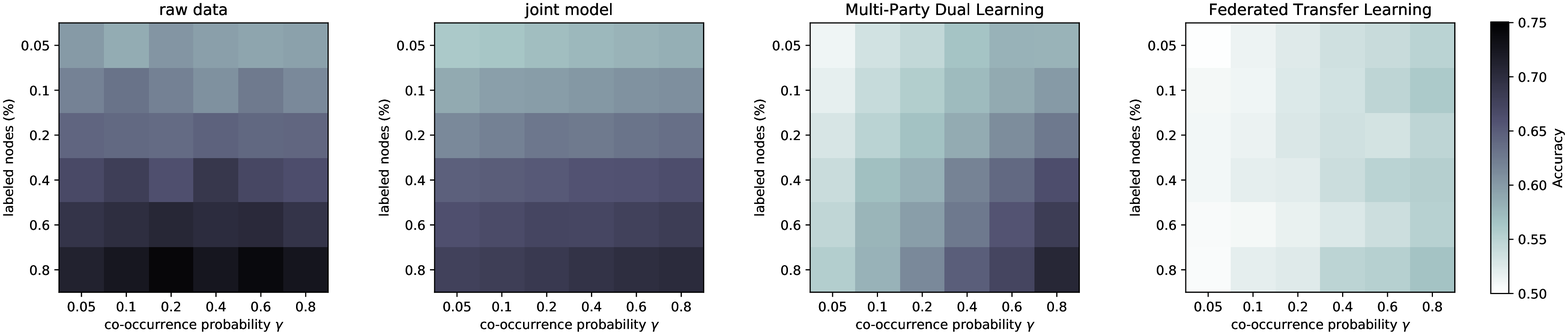}
\caption{Accuracy of node classification with different labeled nodes and co-occurrence probabilities on Google+.}
\label{fig:heat1}
\end{figure*}

\begin{figure*}[!t]
  \centering
  \includegraphics[width=0.9\linewidth]{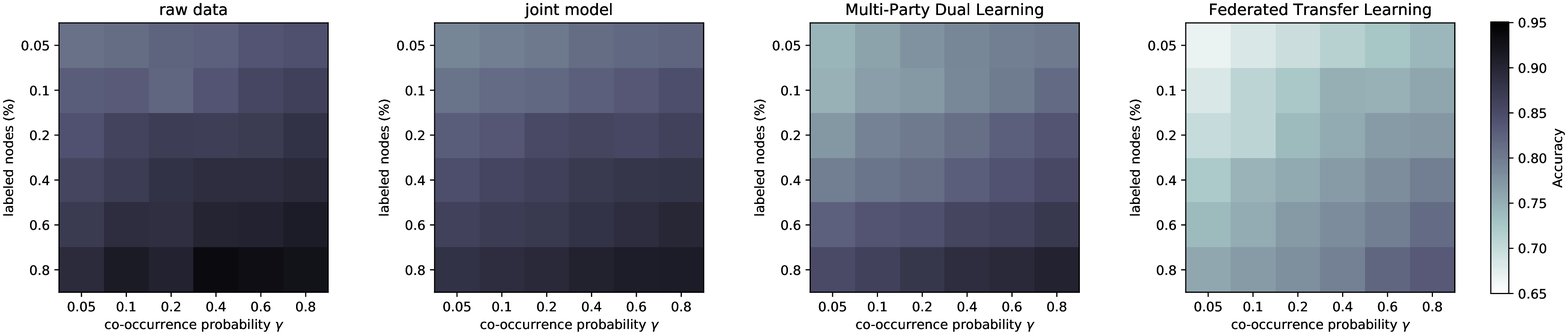}
\caption{Accuracy of node classification with different labeled nodes and co-occurrence probabilities on Hamilton.}
\label{fig:heat2}
\end{figure*}

where representation matrix is the multiplication of adjacency matrix $M_{adj}$ and feature matrix $M_{feat}$, and the missing parts in $M_{adj}$ are set to 0. The process is supplemented by confusion matrices. Without losing generality, for a partitioned representation matrix $M_{m\times f}^P$, assuming it is calculated by the multiplication of a partitioned adjacency matrix $M_{m\times n}^A$ from party $A$ and a partitioned feature matrix $M_{n\times f}^{B}$ from party $B$, which are private data for participants and cannot be transmitted to the third-party collaborator directly. Note that $n$ equals to $N$ and we have $n>m$. Party $B$ postmultiplies $M_{n\times f}^{B}$ by a confusion matrix $M^c_{f\times f}$ and sends $M_{n\times f}^{B}M^c_{f\times f}$ to party $A$, then party $A$ premultiplies it by $M_{m\times n}^A$ and send $M_{m\times n}^AM_{n\times f}^{B}M^c_{f\times f}$ back to party $B$, which calculates node representations as follows:
\begin{equation}
M_{rep}=M_{m\times n}^AM_{n\times f}^{B}(M^c_{f\times f}M^{c^{-1}}_{f\times f}).
\end{equation}

Since the rank of each partitioned matrix equals to that of its augmented matrix and both of them are less than $n$, it is impossible to extrapolate links in party $A$ from the node representations $M_{m\times n}^AM_{n\times f}^{B}$ in party $B$. We evaluate the quality of the inferred feature matrix on the link prediction task and that of node representations on the node classification task. Moreover, we test the joint model on both raw data without missing part and that with incomplete adjacency matrix, named ``raw'' and ``joint'', respectively, to validate the effect of links connecting the two parties. For the MPDL and FTL model, experiments are conducted with private data to simulate the real two-party machine learning problem.

To predict whether there are links in the missing parts in $M_{adj}$, we adopt a standard evaluation metric Area Under Curve score (AUC), which indicates the probability that the potentially connected nodes are more similar than irrelevant ones. It is observed in Fig.~\ref{fig:link} that the joint model with raw data achieves superior performance, and the incomplete adjacency matrix leads to a decrease in the AUC score. The performance of our MPDL model improves as we increase the co-occurrence probability $\gamma$, and it even outperforms the joint model with raw data when $\gamma=0.8$ on Google+, which means the proposed approach is capable of exploring and enhancing the intrinsic probabilistic connection of data. The FTL model has a poor performance on the link prediction task, as the mapped feature vectors weaken the original similarity among features of nodes.

The classification experimental results for Google+ and Hamilton are shown in Fig.~\ref{fig:heat1} and Fig.~\ref{fig:heat2}, where darker colors represent higher classification accuracy. We also evaluate the effect of the number of labeled nodes, which also varies from 0.05 to 0.8, and the rest serve as test data. For both datasets, the joint model with full raw data constantly outperforms other models, and missing parts in adjacency matrices and feature matrices result in varying degrees of accuracy loss. Since the raw data are complete and lossless, co-occurrence probability $\gamma$ has no effect on its results, and the joint model with raw data is slightly better than the same model with an incomplete adjacency matrix. Note that, the essential difference between the joint model and MPDL is the completeness of the feature matrix, which further affects the completion of the adjacency matrix, as shown in Fig.~\ref{fig:link}. Therefore, the MPDL and FTL model could have poor performance, especially when the number of features is large (Google+) or $\gamma$ is small. The MPDL model shows competitive performance compared with the joint model on Hamilton, which verifies the ability of our method to infer data in the other party. Moreover, the FTL model has the same trend as MPDL, while it is limited by the bottleneck of information loss in the knowledge transfer process and is inferior to our MPDL model.

\begin{figure*}[!t]
  \centering
  \subfigure{
    \includegraphics[width=0.4\linewidth]{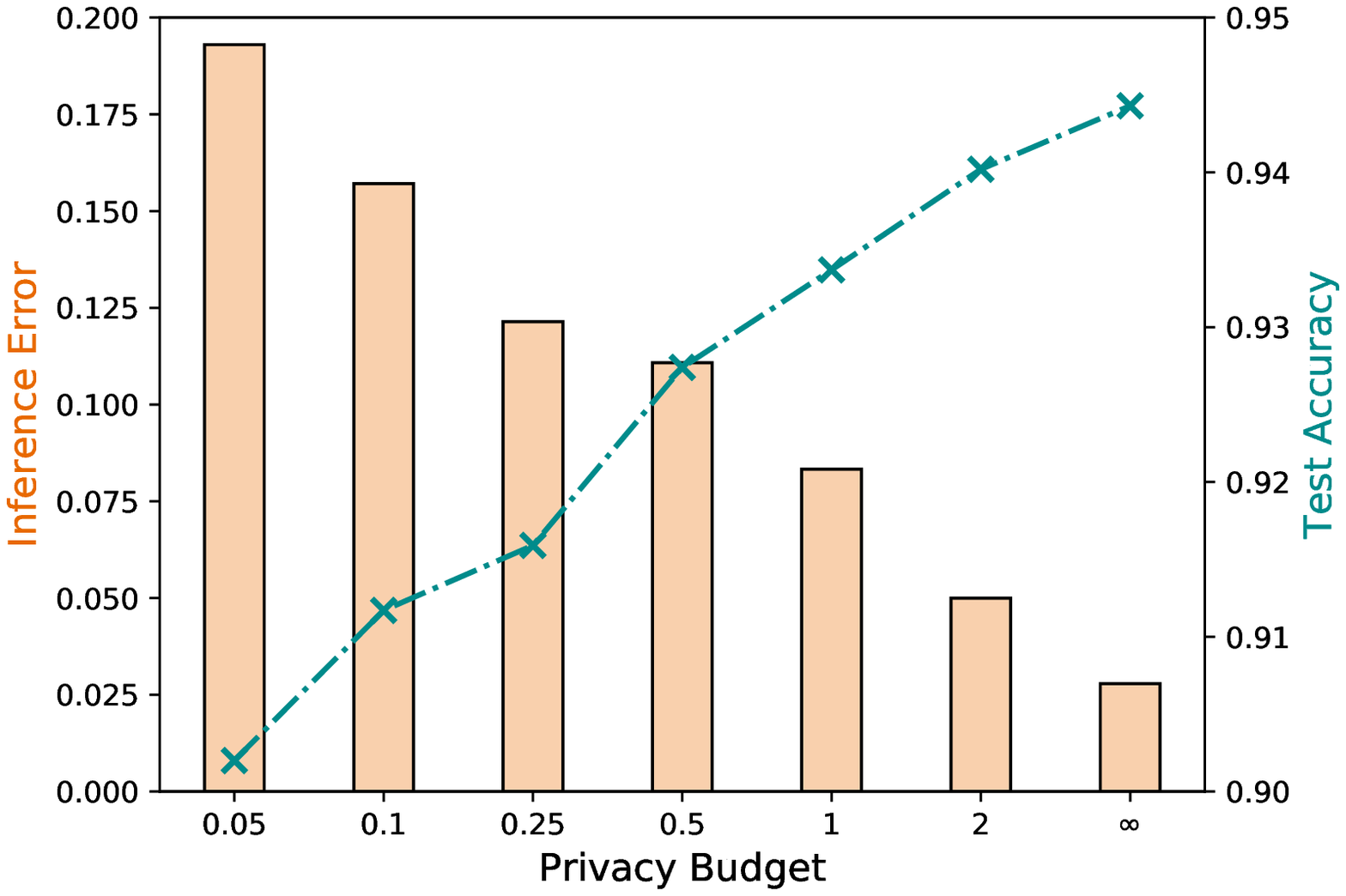}}
    \qquad\qquad
  \subfigure{
    \includegraphics[width=0.4\linewidth]{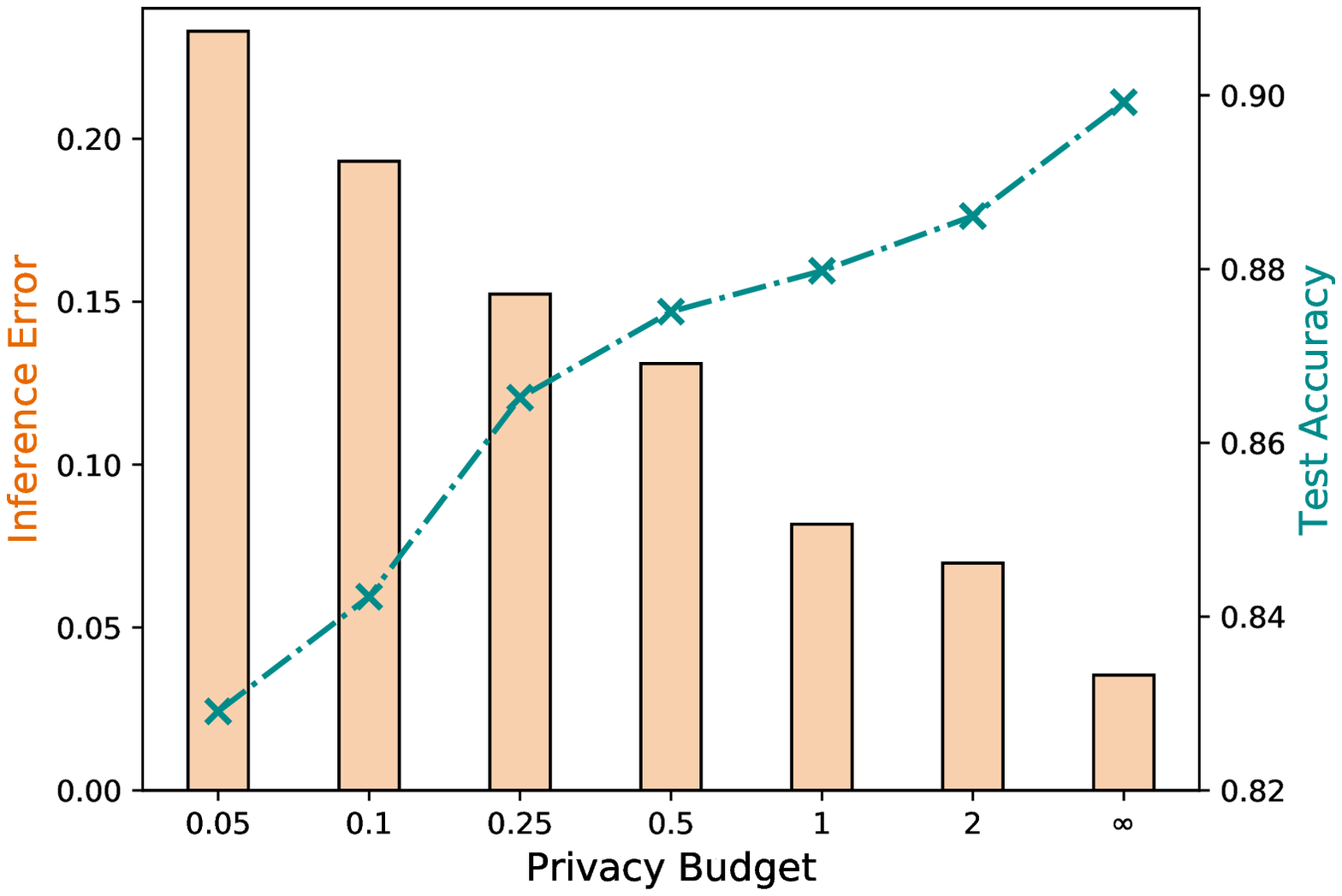}}
  \caption{The impact of the privacy budget $\epsilon$ on the accuracy and inference error on MNIST (Left) and Breast Cancer (Right).}
\label{fig:privacy}
\end{figure*}

\subsection{Privacy Evaluation}
\label{peva}
By introducing dual learning technique in the multi-party learning scheme, the performance of the proposed MPDL framework significantly improves compared with that of a jointly trained central model or the Federated Transfer Learning approach. Nevertheless, the effect of the differential privacy on the test accuracy and the data privacy level is not yet assessed, since the privacy budget $\epsilon$ is fixed in the performance evaluation experiments. Therefore, extensive experiments are conducted on the influence of the differential privacy, and a novel method is proposed to investigate how the privacy budget affects the accuracy and privacy.

According to the data privacy definition in Section~\ref{sec:31}, each participant cannot derive the other's input data in the multi-party dual learning scheme. In order to quantify the level of privacy protection in the dual learning process, the Mean Absolute Error (MAE) between the raw data and the inferred data is calculated, and a smaller error indicates a higher degree of privacy leakage. The equivalence is reasonable since the inference error numerically represents how much private data each participant can derive from the other. Relatively, the loss of dual models is the MAE between the inferred data and the perturbed data, i.e. the output of the affine transformation layer. The experiments are conducted on the MNIST and the Breast Cancer datasets, in which the co-occurrence probability is set to 0.1, and CNN is employed for the former and MLP for the latter. The impact of the privacy budget $\epsilon$ on the central model accuracy and inference error are illustrated in Fig.~\ref{fig:privacy}. When the value of $\epsilon$ is infinity, differential privacy is not employed and the framework retrogrades to a non-private setting. It is important to note that the accuracy and privacy show a negative correlation. Along with the privacy budget cuts, more noise is injected to features of the raw data, and it results in a decrease in data availability and model accuracy. To be specific, the amount of Laplace noise directly affects the MAE of the raw data and perturbed data. Furthermore, the correlation among features is somewhat undermined in the perturbed data, and it further brings negative effect on the performance of both central model and dual models. In other words, the perturbed data serves as a bridge in the calculation of the MAE between the inferred data and the raw data. Through adjusting the privacy budget $\epsilon$, the MPDL framework is able to achieve a balance between the data privacy and the accuracy of central model.

\section{Conclusion}
\label{para:5}
In this work, we develop a novel multi-party dual learning framework and expand the multi-party learning scheme to broader real-world applications. To address the problem of lacking co-occurrence training samples, we introduce privacy-preserving dual learning to generate reliable overlapping pairs. Moreover, encrypted loss and gradients are transmitted among participants without disclosing any information about raw data. The proposed method significantly reduces the dependence on labeled co-occurrence samples, and it is able to achieve superior performance over the non-distributed model with data gathered at one place without privacy constraints. We conduct extensive experiments on several real-world datasets, and results demonstrate that our MPDL model significant outperforms other state-of-the-art baselines.

We will explore the following directions in the future:

(1)	We have validated the effectiveness of MPDL on images and graphs processing, and we will make it scalable for more meaningful tasks such as text processing. Meanwhile, we plan to investigate the feasibility of recurrent neural networks \cite{li2018independently} and graph neural networks \cite{Wu2019Session} \cite{chang2019local} as the central model to improve the scalability of our framework.

(2)	We provide a privacy-preserving general data supplement approach with dual inference, and introduce a feature-oriented differential privacy to preserve private features. In the future work, we try to explore some new schemes to reduce the model complexity and the dependency on the third-party collaborator, thus the method can improve utility and communication efficiency simultaneously.


%





\ifCLASSOPTIONcaptionsoff
  \newpage
\fi


\bibliographystyle{IEEEtran}
\bibliography{ref}
%




\end{document}